\documentclass[10pt,journal,compsoc]{IEEEtran}

\usepackage{graphicx}
\usepackage{booktabs}
\usepackage{multirow}
\usepackage{amsmath}
\usepackage{amsfonts}
\usepackage{amssymb}
\usepackage{pifont}
\usepackage[pagebackref=true,breaklinks=true,letterpaper=true,colorlinks,bookmarks=false]{hyperref}
\hypersetup{colorlinks=true, linkcolor=blue, anchorcolor=blue, citecolor=blue, filecolor=blue, urlcolor=blue} 
\usepackage{subcaption}
\usepackage{color}
\usepackage{xspace}
\newcommand{\eg}{\textit{e.g.}\xspace}
\newcommand{\ie}{\textit{i.e.}\xspace}
\newcommand{\etal}{\textit{et al.}\xspace}
\newcommand{\ours}{LF-Font\xspace}
\newcommand{\methodname}{few-shot font generation with localized style representations and factorization (LF-Font)\xspace}
\usepackage{xcolor}
\definecolor{darkergreen}{RGB}{21, 152, 56}
\definecolor{red2}{RGB}{252, 54, 65}
\newcommand{\yesmark}{\textcolor{darkergreen}{\ding{52}}}
\newcommand{\nomark}{\textcolor{red2}{\ding{56}}}
\newcommand{\myparagraph}[1]{\vspace{2pt}\noindent{\bf #1}}

\newcommand\revision[1]{#1}
\newcommand\addition[1]{#1}

\usepackage{kotex}

\ifCLASSOPTIONcompsoc
  \usepackage[nocompress]{cite}
\else
  \usepackage{cite}
\fi

\ifCLASSINFOpdf

\else

\fi

\begin{document}
\bstctlcite{IEEEexample:BSTcontrol}

\title{Few-shot Font Generation with Weakly Supervised Localized Representations}

\author{Song Park\mbox{*},
	    Sanghyuk Chun\mbox{*},
	    Junbum Cha,
	    Bado Lee,
        Hyunjung Shim%
\IEEEcompsocitemizethanks{
\IEEEcompsocthanksitem Song Park, and Hyungjung Shim are with the School of Integrated Technology, Yonsei University, South Korea.\protect
\IEEEcompsocthanksitem Sanghyuk Chun is with NAVER AI Lab.\protect
\IEEEcompsocthanksitem Sanghyuk Chun, Junbum Cha, and Bado Lee are with the NAVER CLOVA\protect
\IEEEcompsocthanksitem Song Park and Sanghyuk Chun contributed equally to this work. \protect 
\IEEEcompsocthanksitem Hyunjung Shim is a correspondence author (kateshim@yonsei.ac.kr). \protect }
}

\markboth{}
{Park \MakeLowercase{\textit{et al.}}: Few-shot Font generation with Weakly Supervised Localized Representations }

\IEEEtitleabstractindextext{%
\begin{abstract}
Automatic few-shot font generation aims to solve a well-defined, real-world problem because manual font designs are expensive and sensitive to the expertise of designers. Existing methods learn to disentangle style and content elements by developing a universal style representation for each font style. However, this approach limits the model in representing diverse local styles, because it is unsuitable for complicated letter systems, for example, Chinese, whose characters consist of a varying number of components (often called ``radical'') --- with a highly complex structure. In this paper, we propose a novel font generation method that learns localized styles, namely component-wise style representations, instead of universal styles. The proposed style representations enable the synthesis of complex local details in text designs. However, learning component-wise styles solely from a few reference glyphs is infeasible when a target script has a large number of components, for example, over 200 for Chinese. To reduce the number of required reference glyphs, we represent component-wise styles by a product of component and style factors, inspired by low-rank matrix factorization. Owing to the combination of strong representation and a compact factorization strategy, our method shows remarkably better few-shot font generation results (with only eight reference glyphs) than other state-of-the-art methods. Moreover, strong locality supervision, for example, location of each component, skeleton, or strokes, was not utilized. The source code is available at \url{https://github.com/clovaai/lffont} and \url{https://github.com/clovaai/fewshot-font-generation}.
\end{abstract}

}

\maketitle

\IEEEdisplaynontitleabstractindextext
\IEEEpeerreviewmaketitle

\section{Introduction}
\label{sec:intro}
Text is a critical resource of information on the web and publications. Fonts are paints of text-based content design thus fonts have a significant impact on the overall user experience and satisfaction with text-based content; such as logo designs, handouts, magazines, movie posters, and web pages. For example, ``Gotham'' font families are widely adopted for materials that should be objective but convincing (\eg, election handouts, or memorials). On the other hand, ``Comic Sans'' is used for humorous presentations.

However, font design is a labor-intensive and time-consuming work, requiring elaborate handwork by proficient experts, especially for glyph-rich scripts such as Korean and Chinese. For example, when creating a new Korean font library; designers must manually create every single character per font style, where there are a total of 11,172 possible Korean characters (or at least 2,350 widely used characters), while maintaining a coherent font style~\cite{koreantextbook}. For this reason, various font generation methods~\cite{zi2zi, jiang2017dcfont, jiang2019_aaai_scfont, gao2019agisnet, cha2020dmfont} have been investigated to address an automatic font generation problem, which generates a font with a coherent style of the given reference glyph images; the number of reference glyphs varies with the application scenario.

In this study, we addressed a practical font generation scenario: a few-shot font generation problem, in glyph-rich language systems~\cite{sun2018_ijcai_savae,zhang2018_cvpr_emd,gao2019agisnet,liu2019funit,srivatsan2019_emnlp_deepfactorization,cha2020dmfont,lffont,mxfont}, generating a new font library with notably few references ($8$, herein). No additional training procedure (\eg, fine-tuning the model on the reference characters) was performed. We aimed to generate high-quality, diverse styles in the few-shot font generation scenario. The few-shot generation scenario consists of the training and generation stages. During model training, we rely on paired data that are easily accessible by public font libraries. In contrast, at the generation stage, we use only few-shot examples as unseen style references and require no additional model fine-tuning. This scenario is particularly effective in the following application scenarios: 1) when the target-style glyphs are expensive to collect (\eg, historical handwriting), but there is a large database for existing fonts; or 2) computing resources are limited to run additional fine-tuning (\eg, on mobile devices). A popular strategy to tackle this problem is to separate style and content representations from the given glyph images~\cite{sun2018_ijcai_savae,zhang2018_cvpr_emd,gao2019agisnet,srivatsan2019_emnlp_deepfactorization}. These methods generate a full-font library by combining target-style representation and source-content representations.

\begin{figure}
    \centering
    \includegraphics[width=\linewidth]{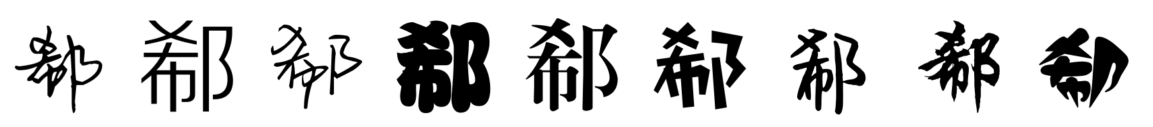}
    \caption{\small {\bf Example font styles of the same character.} Font styles are defined locally by diverse characteristics, such as local strokes or size of components.}
    \label{fig:various_styles}
\end{figure}

A major challenge in font generation tasks is that the font style is often defined locally, \eg, local strokes, serif-ness, or size of sub-characters as shown in Figure \ref{fig:various_styles}. Therefore, a few-shot font generation method should be capable of extracting complex local features from very few reference glyphs. However, previous few-shot font generation methods~\cite{sun2018_ijcai_savae,zhang2018_cvpr_emd,gao2019agisnet} learned and extracted a {\it universal} style representation for each style, which is limited in representing diverse local styles. This is particularly problematic when generating fonts for glyph-rich scripts, such as Chinese or Korean. For example, every Chinese character can be divided into a number of sub-characters or components, with their own meanings, as shown in Figure~\ref{fig:example_annotation}. Hence, the visual quality of a Chinese character tends to be highly sensitive to local damage or a distinctive local component-wise style. As a result, it is difficult to represent highly complex and diverse local styles using only a universal style representation.  

Cha~\etal~\cite{cha2020dmfont} reported that many previous methods often fail to transfer unseen styles for the few-shot Korean generation. To alleviate this problem, they proposed a novel architecture named DM-Font. DM-font extracts component-wise local features for all components and utilizes them for generation. Despite its notable generation quality, DM-Font is restricted to {\it complete compositional scripts}, such as Korean and Thai~\cite{cha2020dmfont,cha2020dmfontw}. While each Korean character can be decomposed into a fixed number of components and positions, more complex scripts (\eg, Chinese) can be decomposed into varying components and positions. Consequently, we empirically observe that DM-Font fails to disentangle complex glyph structures and diverse local styles in the Chinese generation task. Furthermore, DM-Font requires that all components are shown in the reference set at least once to construct their memories. In our experiments, DM-Font tends to require more reference images than others; the fonts generated by DM-Font show worse visual quality than other methods using the same number of reference sets (see Figure~\ref{fig:numref_performance}). These drawbacks limit the applicability of DM-Font to generate Chinese characters, consisting of hundreds of components, with a few references.

In this paper, we propose a novel \methodname that utilizes compositionality, a language-specific characteristic, and a weak supervised framework for few-shot font generation. We focus on compositional scripts whose character is decomposed into a number of sub-characters or components, as illustrated in Figure~\ref{fig:example_annotation}. Here, the component labels are weak supervision; the components in the given glyph are known but their locations are unknown. With the component labels of the given glyph as weak supervision, \ours learns to disentangle complex glyph structures and {\it localized} style representations, instead of {\it universal} style representations. Owing to powerful representations, \ours can capture local details in rich text design, thus successfully handling Chinese compositionality. We show that without pixel-level guidance of each component, the proposed method successfully deals with localized representations, resulting in generalizability to novel styles with only very few references compared to previous state-of-the-art font generation methods.

We define the localized style representation as a character-wise style feature that considers both a complex character structure and local styles. Because handling all characters in the glyph-rich script (\eg $>$ 50,000 for Chinese script) is infeasible, we denote {\it the localized style representation} as a combination of component-wise local style representations to reduce the number of required style features (\eg, Chinese script has a few hundred components) (\S~\ref{subsec:localized_style_representation}). However, this strategy can have an inherent limitation: the reference set must cover the entire component set to construct the complete font library. It is infeasible when a target script has a large number of components, \eg, over 200 for Chinese. To solve this issue, we introduce {\it a factorization module}, which factorizes a localized style feature to a component factor and a style factor (\S~\ref{subsec:factorization}). Consequently, our method can generate the whole vocabulary without having the entire components in the reference style, or utilizing strong locality supervision, for example, the location of each component or skeleton.

We demonstrate the effectiveness of the proposed \ours on the Chinese and Korean few-shot font generation scenarios when the number of references is extremely small (i.e., $8$) (\S~\ref{sec:experiment}). Our method significantly outperforms five state-of-the-art few-shot font generation methods with various evaluation metrics. Careful ablation studies on our design choice show that the proposed localized style representation and factorization modules are an effective choice to tackle our target problem successively.

This work is an extensive version of our AAAI 2021~\cite{lffont}. Compared to the AAAI 2021 work, this paper includes the following additional contributions: (a) reformulation of the original localized features using a weakly supervised learning problem (\S\ref{sec:intro}, \S\ref{sec:method}), and related discussions (\S\ref{sec:relwork}); (b) additional analyses and ablation studies to demonstrate the effectiveness of the proposed weakly supervised localized style representations; (c) additional comparisons by varying the reference size from one-shot to a many-shot; (d) extension to few-shot Korean generation; (e) removing component conditions in test-time, resulting in showing superior few-shot generation performances on unseen languages (e.g., Chinese to Korean generation); and (f) showing the effectiveness of few-shot font generation methods to the character recognition systems.

\section{Related Works}
\label{sec:relwork}

\myparagraph{Font generation as image-to-image translation and style transfer.}
Image-to-image (I2I) translation~\cite{isola2017_cvpr_pix2pix, zhu2017_iccv_cyclegan} aims to learn a mapping between source and target domains while preserving the contents in the source domain, for example, day to night. Recent I2I translation methods are extended to learn a mapping between multiple diverse domains~\cite{stargan, liu2018unified, yu2019multi, starganv2} (\ie, multi-domain translation); thus, they can be naturally adopted into the font generation problem. For example, \cite{zi2zi} attempted to solve the font generation task via paired I2I translation by mapping a fixed ``source'' font to the target font. Inspired by this approach, several recent papers~\cite{zi2zi, jiang2019_aaai_scfont, gao2020_aaai_chirogan, huang2020_eccv_rdgan, wu2020calligan} addressed the font generation task by presuming a large set of references, \eg, 775~\cite{jiang2019_aaai_scfont}, and additional finetuning for generating each font. Our scenario focuses on the few-shot font generation task which requires only a few reference glyphs without any finetuning.

\myparagraph{Font generation as a content-style disentanglement.}
Our application scenario is also related to content-style disentanglement approaches, such as style transfer. Style transfer methods use a pre-trained model that can represent content and style representations separately by their purpose, \eg, artistic styles such as texture or painting styles~\cite{gatys2016neuralstyle, adain, wct} or photorealistic styles such as diverse lighting or materials~\cite{deepphotostyle, photowct, wct2}. However, style transfer methods are designed to capture the ``style'' as global artistic textures or photorealistic colors of the given images, while the font style is defined locally. For this reason, the na\"ive extension of existing style transfer approaches to the font generation task is not effective. The unique characteristics of the font domain motivate the design of a font domain-specific approach for few-shot font generation tasks.

Similarly, other content-style disentanglement approaches, such as image-to-image translation tasks, suffer from the same issue. For example, most of image-to-image translation methods to extract style information from the references \cite{huang2018munit, liu2019funit, starganv2, karras2019style, karras2020analyzing} relies on AdaIN \cite{adain}, designed for style transfer. However, as we observed in our experiments with FUNIT \cite{liu2019funit}, a universal style extractor based on AdaIN cannot capture the complex local styles of glyphs. In this paper, we propose a component-wise localized style representation to mitigate the issue by utilizing component labels as a weak supervision.

\myparagraph{Attribute-conditioned generation and font generation tasks.}
Attribute-conditioned (AC) generation methods aim to generate an image by given attribute conditions, such as ``brown hair color'', ``female'' \cite{yan2016attribute2image, choe2017face, lu2018attribute, he2019attgan}. Since a glyph is conditioned by many components, font generation tasks can be viewed as AC generation tasks. However, there are significant differences between AC generation tasks and font generation tasks. First, a glyph is uniquely defined for each font design, while an attribute-conditioned image can be mapped to various images by different viewpoint, background or lighting \cite{yan2016attribute2image}. For example, a facial image can be mapped to other facial images while keeping identity by changing identity independent information such as viewpoint or background. On the other hand, if we change a component condition of the given glyph, the meaning of the glyph will be no longer preserved. Second, in the font domain, it is easy to obtain glyph images with the same content but different styles, while in general image domains, \eg, facial images, it is impossible to collect all possible pairs for the given attributes. In this paper, we focus on font-specific domain knowledge such as compositionality to capture complex local styles of glyph images.

\myparagraph{Few-shot font generation.}
The few-shot font generation task aims to generate new glyphs with very few style references without additional fine-tuning. The mainstream of few-shot font generation attempts to disentangle content and style representations, specialized in font generation tasks. For example, AGIS-Net~\cite{gao2019agisnet} proposed a font-specialized local texture discriminator and local texture refinement loss. Unlike other methods, DM-Font~\cite{cha2020dmfont} disassembles glyphs into stylized components and reassembles them into new glyphs by utilizing a strong compositionality prior. DG-Font \cite{xie2021dgfont} utilizes deformable convolution \cite{zhu2019deformable} to unsupervised font generation tasks, \ie, assuming there is no paired glyphs across different styles. In our scenario, we aim to generate standard true type font libraries hence we assume that there exist a number of paired glyph images across different styles by rendering images from the existing font libraries.

Despite notable improvements over the past few years, previous few-shot font generation methods have significant drawbacks. They are 1) infeasible to generate complex glyph-rich scripts~\cite{azadi2018mcgan}; 2) fail to capture the local diverse styles~\cite{sun2018_ijcai_savae,zhang2018_cvpr_emd,gao2019agisnet,liu2019funit,srivatsan2019_emnlp_deepfactorization}; or 3) loss of complex content structures~\cite{cha2020dmfont,cha2020dmfontw}. Our method employs localized style features trained by weak component supervision to capture the local diverse styles. To prevent the generated glyphs from losing the complex content structure, we used a content encoder to preserve the content structure. As a result, the samples generated by our method show high visual quality for complex glyph-rich scripts, \eg, Chinese.

\myparagraph{Many-shot font generation methods.}
Although we only focused on the few-shot font generation problem, several papers have addressed the Chinese font generation task with numerous references or additional fine-tuning. SCFont~\cite{jiang2019_aaai_scfont} and ChiroGAN~\cite{gao2020_aaai_chirogan} extracted a skeleton or stroke from the source glyphs and translated it into the target style. They required a large number of references for generating glyphs with a new style using the I2I framework, \eg, $775$~\cite{jiang2019_aaai_scfont}. Instead of expensive skeleton or stroke annotations, different approaches~\cite{sun2018_ijcai_savae,huang2020_eccv_rdgan,wu2020calligan,cha2020dmfont} utilize the compositionality to reduce the expensive search space in the character space to smaller component space. For example, RD-GAN~\cite{huang2020_eccv_rdgan} employs an additional LSTM architecture~\cite{lstm} to capture the compositionality of the given glyph. However, RD-GAN is designed to reconstruct unseen characters in a fixed style; thus, it cannot be applied to our few-shot generation scenario, which aims to generate characters with unseen styles. CalliGAN~\cite{wu2020calligan} encodes the styles by one-hot vectors; thus, it requires additional fine-tuning to create an unseen style during the training. ChiroGAN~\cite{gao2020_aaai_chirogan} aims to solve unpaired font generation tasks as unpaired image-to-image translation tasks~\cite{zhu2017_iccv_cyclegan}. However, in our scenario, glyph images can be easily rendered from an existing font library, as building a paired training dataset is cheap and does not limit practical usage. Similarly, StrokeGAN \cite{zeng2021strokegan} employs one-bit stroke encoding to capture the key mode information of Chinese characters on unpaired font generation tasks. Both StrokeGAN and our method utilize the compositionality (or stroke information), but StrokeGAN only uses the primal strokes while ours utilizes the component labels where strokes and components correspond to ``character bytes'' and ``tokenized words'' in natural language processing. We did not compare our method to many-shot font generation methods because their methods are not applicable to our scenario: there exists very few references (\eg, 8 in our experiments) and no finetuning procedure is allowed.

\myparagraph{Weakly supervised object recognition.}
In this study, we utilized weak component-level supervision to learn localized features. Our weak supervision is image-level multi-labels without pixel-level annotations, that is, the exact position of each component is unknown. A similar scenario is widely adopted in many weakly supervised vision recognition tasks; such as weakly supervised object localization~\cite{choe2020wsoleval, choe2020wsoleval_extension}, weakly supervised object detection~\cite{bilen2016weakly}, or weakly supervised semantic segmentation~\cite{papandreou2015weakly, xu2015learning}. These weakly supervised vision recognition techniques have shown that with only image-level weak supervision, they can achieve a reasonable localization ability, for example, locating the target objects in the image. However, these techniques are specifically designed for object recognition tasks by introducing a task-specific module design and training strategy. Hence, it is difficult to apply their schemes directly to the few-shot font generation problem. For example, object localization techniques commonly suffer from performance bias; the estimated localization map captures only the most discriminative regions of the object. To resolve this bias, existing methods have been developed to expand the localization map. For this purpose, adversarial complementary learning (ACoL)~\cite{zhang2018adversarial} for weakly supervised object localization proposes a two-head architecture --- where one head acts as an adversary, and attempts to erase the high-score region produced by the other. The attention-dropout layer (ADL)~\cite{choe2019attention} for the same problem is proposed to erase the highly attended region by channel attention. Both techniques propose a task-specific module for expanding the localization maps and require a generic-purpose vision backbone; namely the ImageNet-pretrained network. Hence, despite the advances of previous methods in weakly supervised object recognition, a new methodology is required to solve a few-shot generation task for a specific visual domain, that is, the font domain.

\section{Few-shot Font Generation with Weakly supervised Localized Representations}
\label{sec:method}

We propose a novel few-shot font generation framework, \methodname. In this section, we introduce component-wise localized features trained by language-specific weak component labels.

\begin{figure}[t]
    \centering
    \includegraphics[width=0.6\linewidth]{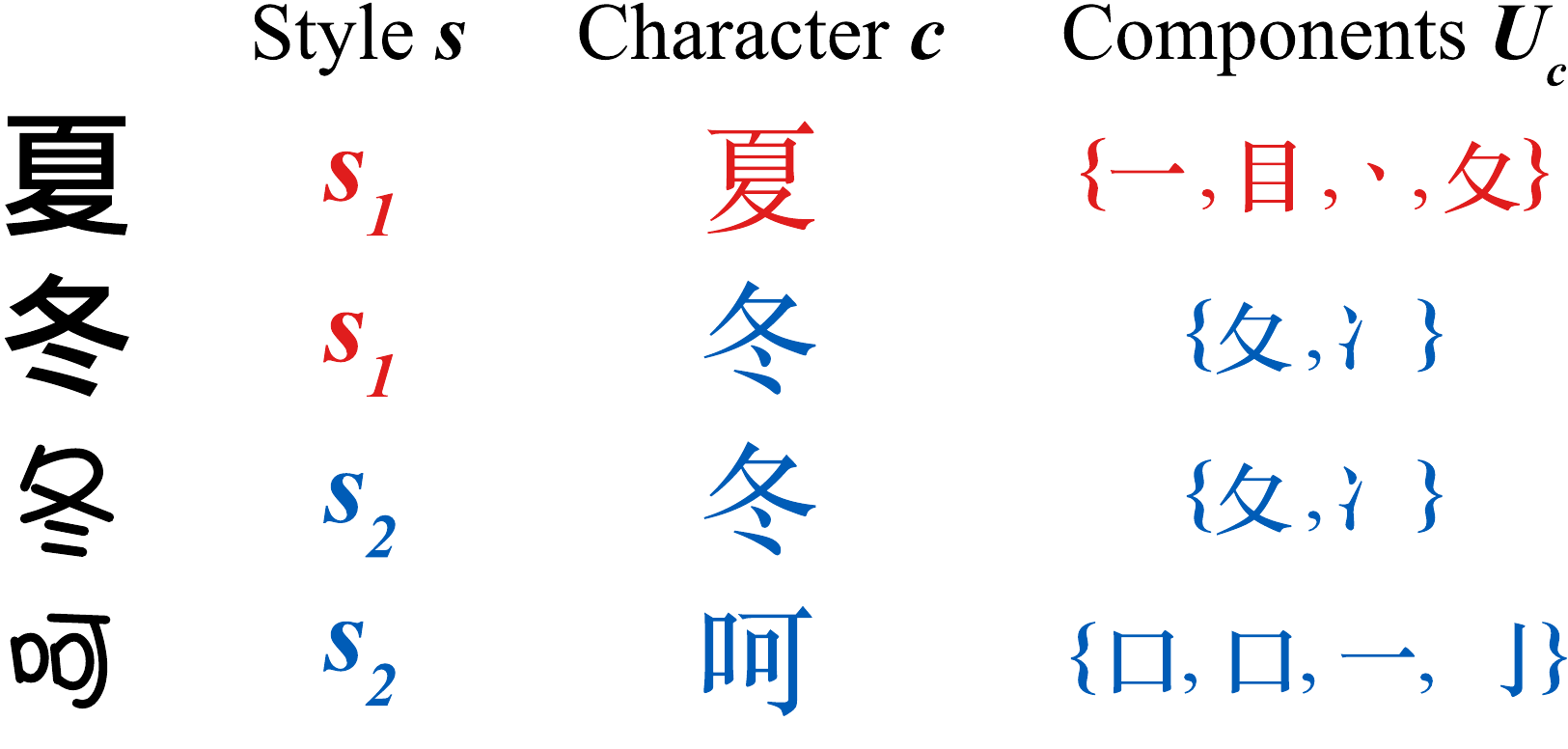}%
    \caption{\small {\bf Annotation examples.} The character label $c$, style label $s \in \{s_1, s_2\}$, and the component label set $U_c$ are shown.}%
    \label{fig:example_annotation}
\end{figure}

\begin{figure}[t]
    \centering
    \includegraphics[width=0.65\linewidth]{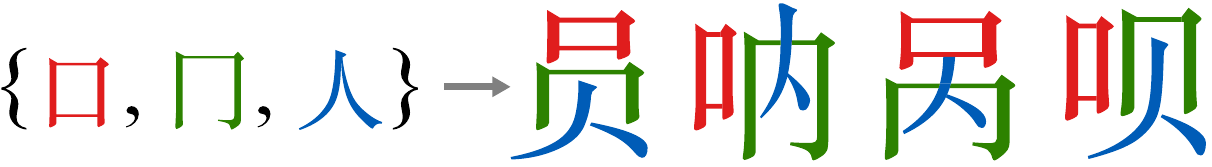}%
    \caption{\small {\bf Characters from the same component set.} Examples show that a component set is mapped to diverse characters.}%
    \label{fig:twins}
\end{figure}

\subsection{Compositionality: a language-specific image-level weak supervision}
\label{subsec:compositionality}

A major challenge in the font generation task for a glyph-rich script, such as Chinese ($>$ 50K glyphs) or Korean ($\approx$ 11K glyphs) is the large number of characters to generate. Many letter systems have a language-specific property, {\it compositionality}; a character can be decomposed into a number of sub-characters or components. It is worth noting that among the top 30 popular letter systems, 24 have compositionality, \eg, Chinese, Hindi, Arabic, Japanese, Korean, and Thai (--- See Appendix for examples). Utilizing the predefined decomposition rule, all characters can be represented by only a small number of components, for example, 68 components for Korean and about a few hundred components (371 in our experiments) for Chinese\footnote{We use the character decomposition data from Wikimedia Commons for Chinese decomposition. \url{https://commons.wikimedia.org/wiki/Commons:Chinese_characters_decomposition}}. In Figure~\ref{fig:example_annotation}, we illustrate the example component labels and other annotations of Chinese characters. As seen in the bottom example of this figure, some Chinese characters contain duplicated components ~\ref{fig:example_annotation} (``口'' is duplicated in this example). We used all duplicated components as the input for the proposed method.

The component labels are weak supervision; while the components in the given glyph are known, their locations are unknown. Furthermore, a component label set can be mapped to multiple characters. Figure~\ref{fig:twins} indicates that by combining three components, four different characters can be rendered, where the detailed shape of the component depends on the component location. For example, ``口'' (red component in the figure) shows different width and height ratios, and variable sizes upon different locations. We designed our model to capture local component features and to correctly combine them to express the structure of the target character. In the next subsections, we describe how \ours can deal with both localized component-wise features and the global structure of the target character.

\subsection{Problem definition}
\label{subsec:probdef}

We define three annotations for a glyph image $x$: the style label $s \in \mathcal S$, the character label $c \in \mathcal C$, and the component labels $U_c = [u^c_1, \ldots, u^c_m]$, where $m$ is the number of components in character $c$. Here, each character $c$ can be decomposed into components $U_c$ using the predefined decomposition rule, as shown in Figure~\ref{fig:example_annotation}. In our Chinese generation tasks; the number of styles $|\mathcal S| = 482$, the number of characters $|\mathcal C| = 19,514$, and the number of components $|U| = 371$. In other words, all $19,514$ characters can be represented by a combination of $371$ components. Our problem definition is not limited to Chinese, but is easily extended to other languages, as shown in \S\ref{subsec:more-lang}.

\begin{figure*}[t]
    \centering
    \includegraphics[width=\linewidth]{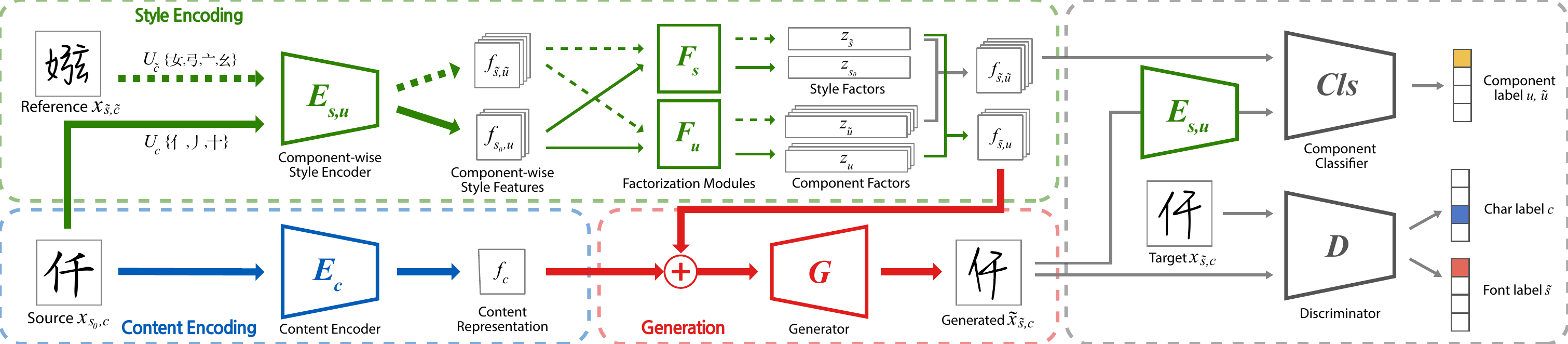}%
    \caption{\small {\bf Overview of \ours.} \ours consists of four parts; the content-encoding $E_c$, the style-encoding $E_{s,u}, F_s, F_u$, the generation $G$, and the shared modules $D, Cls$ for training. $E_c$ encodes the source glyph to the content representation $f_c$. The source (solid line) and reference (dashed line) images are encoded to component-wise style features $f_{s,u}$, and further factorized into style and component factors $z_s, z_u$. $z_s, z_u$ are combined to the character-wise style representation $f_{s,c}$ of the target glyph. The generator $G$ synthesizes the target glyph from the $f_c$ and $f_{s,c}$.
 }%
    \label{fig:architecture}
\end{figure*}

The goal of the few-shot font generation task is to generate a glyph $x_{\widetilde s, c}$ with unseen target styles $\widetilde s$ for all $c \in \mathcal C$ with very few references $x_{\widetilde s, \widetilde c} \in \mathcal X_r$, \eg, $| \mathcal X_r | = 8$. A common framework for few-shot font generation is to learn a generator $G$ which takes the style representation $f_{\widetilde s} \in \mathbb R^d$ from $\mathcal X_r$ and the content representation $f_c \in \mathbb R^d$ as inputs. Then synthesizing a glyph $x$ having the reference styles $\widetilde s$, but representing a source character $c$. Formally, a few-shot font generation task can be represented as follows:
\begin{align}
\label{eq:fewshotfontgeneral}
\begin{split}
    &x_{\widetilde s, c} = G( f_{\widetilde s}, f_c ), \\
    f_{\widetilde s} = &E_s(\mathcal X_r) ~\text{and}~ f_c = E_c( x_{s_0, c} ),
\end{split}
\end{align}
where $s_0$ is the source style label.

\subsection{Localized style representations}
\label{subsec:localized_style_representation}
Previous methods assume that the style representation $f_s$ is universal for each style $s$. However, the universal style assumption can overlook complex local styles, resulting in poor performance for unseen styles as pointed out by~\cite{cha2020dmfont}. Here, we design the style encoder $E_s$ to encode a character-wise style. This strategy is useful when a style is defined locally and diversely as Chinese characters. However, the large vocabulary size of Chinese script ($|\mathcal C| > 20,000$) makes it impossible to exploit all character-wise styles.

Instead of handling all character-wise styles, we first represent the character as a combination of multiple components, and develop component-wise styles to minimize redundancy in character-level representations. For this, we utilize the component set $U_c$ instead of the character label $c$, where $|U| \ll |\mathcal C|$ (371 and 19,514 in our experiments). We extract a component-wise style feature $f_{s,u}(x, u) = E_{s,u}(x, u) \in \mathbb R^d$ from a reference glyph image $x$, and a component label $u \in U_c$ --- by introducing a component-wise style encoder $E_{s,u}$. Here, the component labels are image-level weak supervision; we only know that there exists the given components but we do not know where are they. Our image-level component weak supervision is similar to the weak supervision adapted by weakly-supervised object localization (WSOL), weakly-supervised object detection (WSOD), or weakly-supervised semantic segmentation (WSSS). Our goal is to utilize weak image-level component labels to learn localized style representations without expensive pixel-level guidance of each component. From this motivation, we compute {\it the character-aware localized style feature} $f_{s,c}$ by taking the summation over component-wise features $f_{s,u}$. Now, we can rewrite Eq~\eqref{eq:fewshotfontgeneral} with the proposed character-aware localized style features as follows:
\begin{align}
\label{eq:fewshotfontdm}
\begin{split}
    x (\widetilde s, c) &= G(f_{\widetilde s,c}, f_c ), \quad f_c = E_c( x_{s_0, c} ), \\
    f_{\widetilde s,c} &= \sum_{u \in U_c} f_{\widetilde s,u} = \sum_{u \in U_c} E_{s,u}(x_{\widetilde s, \widetilde c_u}, u),
\end{split}
\end{align}
where $x_{\widetilde s, \widetilde c_u}$ is a glyph image from the reference set $\mathcal X_r$, whose character is $\widetilde c_u$, which contains component $u$. However, this approach still has a significant drawback. Because we require the entire set of component-wise style representations ($|U| = 371$ for Chinese) to reconstruct the whole character set, the component labels of a reference set $\mathcal X_r$ should cover all components. In the Chinese font generation scenario, at least 229 reference characters with a coherent style are required to derive complete component-wise style representations. In the following section, we introduce our solution to reduce the required number of reference images. 

\subsection{Completing missing localized style representations by factorization modules}
\label{subsec:factorization}

Our localized style representation is defined by 1) different styles and 2) a character-wise manner, \ie, a localized style representation $f_{s,c}$ is defined by a character $c$ and a style $s$. As we defined in Eq \eqref{eq:fewshotfontgeneral}, when synthesizing a new character with style $s$ and character $c'$, we need a feature $f_{s,c'}$. Since during test-time, we do not have $c'$ with style $s$, we decompose $f_{s,c}$ into $\sum_u f_{s,u}$ where $u$ is the component of $c$ as illustrated in Eq \eqref{eq:fewshotfontdm}. Our component-wise decomposition has two benefits: 1) every Chinese character can be decomposed into the pre-defined components, therefore we can define $f_{s,c}$ for any character without observing character $c$ with style $s$, 2) the number of representations to learn is dramatically decreased because the number of characters (about 20K in our experiments) is much larger than the number of components (371 in our experiments).

In our scenario, only partial components are observable from the reference set, whereas the other components are not accessible by $E_{s,u}$. However, our formulation in Eq \eqref{eq:fewshotfontdm} needs a full access to the whole components for a novel style. Hence, the localized style feature $f_{s,c}$ for a style $s$ and a character $c$ with unseen components cannot be computed, and therefore, $G$ cannot generate a glyph with $c$. For example, to generate a new font library, one has to prepare at least 229 characters whose components can cover the whole 371 components, while our goal is few-shot font generation with very few references (\eg, 8).

To tackle this problem, we formulate the few-shot font generation problem as a reconstruction problem. Here, we treat each component-style combination as an entry of the matrix and we assume that some component-style entries are missed. Our goal is to reconstruct a novel combination of content-style with previously observed data entries. We employ the factorization modules motivated by the low-rank matrix factorization (MF) that assumes the data matrix $X$ has low rank $k$, \ie, $X = A^\top B$ where $\text{rank}(A) = \text{rank}(B) = k << \text{rank}(X)$. With this assumption, the value of the entry $(i, j)$ can be computed by $a_i^\top b_j$. However, applying conventional MF algorithms to every novel style is inefficient and computationally inefficient. Inspired by classical matrix completion approaches~\cite{candes2009exact,cai2010singular}, we decompose the component-wise style feature $f_{s,u} \in \mathbb R^d$ into two factors: a component factor $z_u \in \mathbb R^{k \times d}$ and a style factor $z_s \in \mathbb R^{k \times d}$, where $k$ is the dimension of the factors. Formally, we decompose $f_{s,u}$ into $z_s$ and $z_u$ as follows:
\begin{equation}
\label{eq:factorization}
    f_{s,u} = \boldsymbol{1}^\top (z_s \odot z_u),
\end{equation}
where $\odot$ is an element-wise matrix multiplication, and $\boldsymbol{1} \in \mathbb R^k$ is an all-ones vector. Eq~\eqref{eq:factorization} can be interpreted as the element-wise matrix factorization of $f_{s,u}$. In practice, we extract the style factor $z_s$ from the reference set and combine them with the component factor $z_u$ from the source glyph to reconstruct a component-wise style feature $f_{s,u}$ for the given source character $c$. Notably, \cite{tenenbaum2000separating, srivatsan2019_emnlp_deepfactorization} also uses a factorization strategy for font generation; however, they directly apply factorization to the complex glyph space (\ie, each element is an image), while \ours factorizes the localized style features into the style and the content factors.

Traditional matrix completion methods require heavy computations and memory consumption. For example, expensive convex optimization~\cite{candes2009exact}, or alternative algorithms ~\cite{cai2010singular} are infeasible in our scenario by repeatedly applying matrix factorization $d$ times to obtain a $d$-dimensional feature $f_{s,u}$. Instead, we propose a style and component factorization module $F_s$ and $F_u$ that extract factors $z_s, z_u \in \mathbb R^{k \times d}$ from the given feature $f_{s,u} \in \mathbb R^d$ as follows:
\begin{equation}
\label{eq:factorization_modules}
    z_s = F_s (f_{s,u}; W, b), \quad z_u = F_u (f_{s,u}; W, b).
\end{equation}
We used a linear weight $W = [w_1; \ldots; w_k] \in \mathbb R^{k \times d}$ and bias $b \in \mathbb R^k$ as a factorization module, where each factor is computed by $z = [w_1 \odot f_{s,u} + b_1; \ldots; w_k \odot f_{s,u} + b_k]$.

Note that solely employing the factorization modules, \ie, Eq~\eqref{eq:factorization_modules}, does not guarantee that factors with the same style (or component) from different glyphs have identical values. For example, without any constraint, a style factor of $s$ extracted by $u$, $F_s(f_{s,u})$, and a style factor of $s$ extracted by $u'$, $F_s(f_{s,u'})$, will have different values, while we assume that each style factor with the same style is identical --- Eq \eqref{eq:factorization}. Hence, we add a consistency loss that enforces the factors to have the same values for the same content or style. Intuitively, it can be obtained by minimizing all pair-wise distances of the factors, \ie, $\min \sum_{u, u'} \| F_s(f_{s,u}) - F_s(f_{s,u'}) \|_2^2$. Since this equation is identical to minimize the sum of distances between each factor and their average, we train the factorization modules $F_s$ and $F_u$ by minimizing the consistency loss $\mathcal L_{consist}$ as follows:
\begin{align}
\begin{split}
\label{eq:loss_factorization_consist}
    \mathcal L_{consist} = \sum_{s \in \mathcal S} \sum_{u \in \mathcal U} &\| F_s(f_{s,u}) - \mu_s \|_2^2 + \| F_u(f_{s,u}) - \mu_u \|_2^2, \\
    \mu_s = \frac{1}{|\mathcal U|}\sum_{u \in \mathcal U} F_s&(f_{s,u}), \quad \mu_u = \frac{1}{|\mathcal S|}\sum_{s \in \mathcal S} F_u(f_{s,u}).
\end{split}
\end{align}
After training $F$, we can extract $z_s$ from even a random single reference glyph. Furthermore, by combining $z_s$ with the content factor $z_u$ from the known source glyph, we can reconstruct the localized style feature $f_{s, c} = \sum_{u \in U_c} f_{s, u}$ even for the unseen component $u$ in the reference set.

\subsection{Generation}
\label{subsec:generation}
Once \ours is trained with many paired training samples, it is able to generate any unseen style fonts with only a few references by extracting the style factor $z_{\widetilde s}$ from the reference glyphs, and by extracting $z_u$ and $f_c$ from the known source glyphs. Then, we combine $z_c$ and $z_{\widetilde s}$ to generate the localized style feature $f_{\widetilde s, u}$, as described in \S\ref{subsec:factorization}. Finally, we generate a glyph $x$ using Equation \eqref{eq:fewshotfontdm}. Formally, \ours consists of three sub-modules, as illustrated in Figure~\ref{fig:architecture}.

\myparagraph{Style encoding.}
\ours encodes the localized style representation $f_{s,c}$ by encoding the component-wise features $f_{s,u}$ as formulated in Eqs~\eqref{eq:fewshotfontdm},~\eqref{eq:factorization} and~\eqref{eq:factorization_modules}. There are three main modules in this stage: the component-wise style encoder $E_{s,u}$, and the style and content factorization modules $F_s$ and $F_c$. $E_{s,u}$ is simply defined by a conditional encoder as previous generative models \cite{starganv2, karras2020analyzing}, where a component label $u$ is used for the condition label, and encodes a glyph image $x$ into several component-wise style features $f_{s,u}$. 

More specifically, the component-wise style encoder $E_{s,u}$ consists of five modules: convolution, residual, component-conditional, global-context~\cite{cao2019gcnet}, and convolutional block attention (CBAM)~\cite{woo2018cbam}. The component-conditional module learns a set of channel-wise biases where each bias value is in charge of each component. The component-wise style encoder reflects the given component condition by adding the corresponding channel-wise bias learned by the component-conditional module to the intermediate features.

A component-wise style feature $f_{s,u}$ is factorized into the style factor $z_s$, and component factor $z_u$, with factorization modules $F_s$ and $F_u$, respectively. We combine the style factor $z_{\widetilde s}$ from the reference glyphs, and component factor $z_u$ from the source glyph, to reconstruct the component-wise feature $f_{\widetilde s,u}$. If there is more than one reference sample, we take the average over the style factors, extracted from each reference glyph, to compute $z_{\widetilde s}$.

Our style encoding requires the explicit component labels due to the component-wise style encoder $E_{s,u}$. Utilizing character labels as weak supervision during training is affordable because we use true-type fonts easily accessible from web for training our models. However, assuming that character labels are available even in test-time can limit the applicability of \ours when the given reference glyph images are unlabeled (\eg, historical handwriting). Furthermore, it hinders \ours to generate characters with unseen components during the training (\eg, generating glyph images for different language systems).

To mitigate the issues, we introduce an auxiliary classifier that predicts the character label of the given glyph image. Our prediction-based inference strategy has two benefits; 1) we can remove the dependency of character labels in test-time. It can be beneficial when the reference characters are unlabeled, e.g., historical handwriting. 2) the prediction-based strategy makes \ours handle characters having unseen components in the training set. For example, the original LF-Font will not be effective if the target character is from a different language system, such as Korean characters. The detailed discussion and experimental results are in \S\ref{subsec:psuedo-character}.

\myparagraph{Content encoding.}
Although our style encoding strategy effectively captures the local component information, it requires guidance on the complex global structure (\eg, relative locations of components) of each character, because a component set can be mapped to many characters (see Figure~\ref{fig:twins}). We employ the content encoder $E_c$ to capture the complex global structural information of the source glyph. It facilitates the generation of the target glyph while preserving complex structural information without any strong localization supervision of the source glyph.

\myparagraph{Generation.}
Finally, the generator $G$ produces the target glyph $\widetilde x_{\widetilde s, c}$ by combining the localized style representations $f_{\widetilde s,c}$ from the style encoding and the global complex structural representation $f_{c}$ from the encoding.

\subsection{Training}
\label{subsec:training}
Given the source glyph $x$ and the references $\mathcal X_r$ with the target style $s$, \ours learns the style encoder $E_{s,u}$, the content encoder $E_c$, the factorization modules $F_s, F_u$, and the generator $G$ to generate glyph $\widetilde x$. We fix the source style $s_0$ during training and optimize the model parameters with diverse reference styles using the following losses:

\myparagraph{Adversarial loss.}
We employ a multi-head conditional discriminator for style label $s$ and character label $c$. The hinge GAN loss~\cite{zhang2019sagan} was used.
\begin{align}
\begin{split}
\label{eq:loss_adv}
    \mathcal L_{adv}^D = &-\mathbb E_{(x, s, c) \sim p_{data}} \min\left(0, -1 + D_{s,c}(x) \right) \\
    &- \mathbb E_{(\widetilde x, s, c) \sim p_{gen}} \min\left(0, -1 - D_{s,c}(\widetilde x) \right)\\
    \mathcal L_{adv}^G = &-\mathbb E_{(\widetilde x, s, c) \sim p_{gen}} D_{s,c}(\widetilde x).
\end{split}
\end{align}
\myparagraph{L1 loss and feature matching loss.}
These objectives enforce the generated glyph $\widetilde x$ to reconstruct the ground truth glyph $x$ at the pixel level and feature level.
\begin{align}
\begin{split}
\label{eq:loss_recon}
    \mathcal L_{l1} &= \mathbb E_{(x, s, c) \sim p_{data}} \left[ \| x - \widetilde x \|_1 \right],\\
    \mathcal L_{feat} &= \mathbb E_{(x, s, c) \sim p_{data}} \left[\sum_{l=1}^L\| D_f^{(l)}(x) - D_f^{(l)}( \widetilde x ) \|_1 \right]
\end{split}
\end{align}
where $L$ is the number of layers in the discriminator $D$, and $D_f^{(l)} (x)$ is the intermediate feature in the $l$-th layer of $D$.

\myparagraph{Component-classification loss.}
We employ an additional component-wise classifier $Cls$ that classifies the component label $u$ of the given component-wise style feature $f_{s,u}$. We optimized the cross-entropy loss (CE) as follows:
\begin{equation}
\label{eq:loss_cls}
    \mathcal L_{cls} = \sum_{\widetilde u \in U_{\widetilde c}} \text{CE}(Cls(f_{s, \widetilde u}), \widetilde u) + \sum_{u \in U_c} \text{CE}(Cls(f_{s,u}), u),
\end{equation}
where $f_{s,\widetilde u}$ and $f_{s,u}$ are extracted from the reference glyph $x_{s, \widetilde c}$, and the generated glyph $\widetilde x_{s, c}$.

\myparagraph{Full objective.}
Finally, we optimize \ours by the following full objective function:
\begin{align}
\begin{split}
    &\min_{\substack{E_c, E_{s,u}, G, \\F_s, F_u, Cls}}\max_D~ \mathcal L_{adv (font)} + \mathcal L_{adv (char)} + \lambda_{L1} \mathcal L_{L1} \\
    &\qquad+ \lambda_{feat} \mathcal L_{feat} + \lambda_{cls} \mathcal L_{cls} + \lambda_{consist} \mathcal L_{consist},
\end{split}
\end{align}
where $\lambda_{L1}, \lambda_{feat}, \lambda_{cls}, \lambda_{rep}$ are hyperparameters that control the effect of each objective. We set $\lambda_{L1} = 1.0$ and $\lambda_{feat} = \lambda_{cls} = \lambda_{rep} = 0.1$ throughout all the experiments.

\begin{figure}[t]
    \centering
    \includegraphics[width=.85\linewidth]{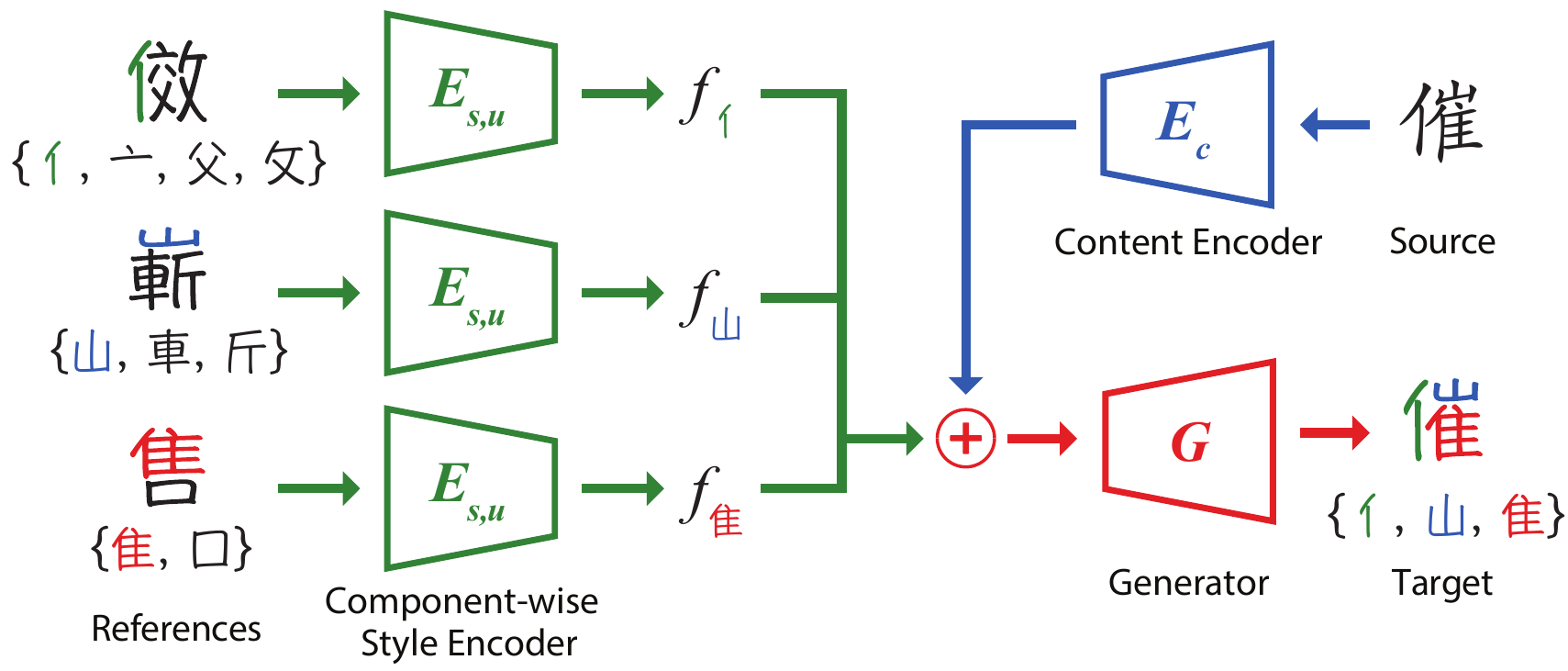}%
    \caption{
    \small {\bf Mini-batch construction for phase 1 training.} During the first phase of the training, we constructed a mini-batch with a coherent style, where the component labels of the input glyphs cover the component labels of the target character.
    }%
    \label{fig:train_batch}
\end{figure}

\myparagraph{Training details.}
We used an Adam~\cite{adam} optimizer with a learning rate of 0.0008 for the discriminator, and 0.0002 for the others. We trained the model in two phases for stability.

Our two-phase training is designed for a stable optimization of factorization modules. The role of the factorization modules is to reconstruct the component-wise style representations for the unseen components in the reference set. We observe that jointly training the factorization module and randomly initialized features at the same time can make the convergence unstable as shown in our ablation study (\S\ref{subsubsec:phase_abl}). Our two-phase training first learns component-wise style representations and then learns the factorization modules. At the first phase, we do not train the factorization modules by building a mini-batch in a way that the target character components are completely covered by the input character components (Figure \ref{fig:train_batch}). After training the component-wise style representations, we train the factorization modules with various styles. The detailed discussion of our two-phase training strategy is in \S\ref{subsubsec:phase_abl}.

In the first phase, we train the model without factorization modules as in Eq~\eqref{eq:fewshotfontdm} until 800k iterations for Chinese and 200k iterations for Korean. Here, the model is trained to generate a target glyph from the component-wise style features $f_{s,u}$, extracted by the style encoder $E_{s,u}$ from the reference set $\mathcal X_r$. The content feature $f_c$ is extracted by the content encoder $E_{c}$ from the source glyph. We constructed a mini-batch with pairs of a reference set, source glyph, and target glyph. To build each pair, we randomly selected a style from the training style set and constructed a reference set and a target glyph, where the components of the target glyph belong to the components in the reference set, but the target glyph is not in $\mathcal X_r$. The source font was fixed throughout the training period. We illustrate an example of mini-batch construction in Figure~\ref{fig:train_batch}. Here, $\lambda_{consist}$ is set to $0.0$, and the generator $G$ and the component classifier $Cls$ take the original component-wise style features from $E_{s,u}$, \eg, without the factorization procedure, as their input.

After a sufficient number of iterations, we began the second phase of training. We jointly trained all modules with the full objective function for 50k iterations. All the component-wise style features used in the first phase are replaced by the reconstructed component-wise style features from style and component factors as Eq \eqref{eq:factorization}. The mini-batch for phase 2 training was constructed differently from phase 1. Our model cannot deal with component-wise style features that are not seen in the reference set if factorization modules are not available. Because the model in the first phase deactivates the factorization modules, we construct the mini-batch by having the images in the reference set and target glyph share the same style. In the second phase, we build the mini-batch by selecting the images with various styles in the reference set, and the target glyph with one of the reference styles. During the second phase of training, we also enforced the model to reconstruct the reference images using the original component-wise style features. This was to construct a model that can handle both the original, and reconstructed component-wise style features.

\subsection{LF-FontMix: Font generation as data augmentation}
\label{subsec:lffontmix}
Mix-based augmentations, such as Mixup \cite{mixup} and CutMix \cite{yun2019cutmix}, are widely used for the state-of-the-art image recognition models. While the simple and random mix strategy surprisingly works well, the augmented images by Mixup and CutMix are unnatural, limiting their power in complex and fine-grained datasets \cite{zhang2021intra}. Character recognition systems are complex and fine-grained systems that also need careful mixing strategies as fine-grained classification tasks.

We propose LF-FontMix, a novel mix augmentation for character recognition tasks. Instead of generating mixed images in the pixel domain, we mix two images in the style factor space. Formally, let $z_{s, 1}$ and $z_{s, 2}$ be style factor of images $x_1$ and $x_2$ defined by Eq \eqref{eq:factorization_modules}, respectively. Then the mixed style feature $\hat f_s$ by LF-FontMix is defined as follows:
\begin{equation}
\label{eq:lffontmix}
    \hat {f_s} = \sum_{u \in U_c} \boldsymbol{1}^\top ( (\lambda z_{s, 1} + (1-\lambda) z_{s, 2}) \odot z_u),
\end{equation}
where $\lambda \in [0, 1]$ is a combination ratio, sampled from $\text{Beta}(\alpha, \alpha)$ as previous methods \cite{mixup, yun2019cutmix}. We generate a mixed image $x_{mix}$ by $x_{mix} (x_1, x_2) = G(\hat f_s, f_c)$. Here, the target label is the same as the target character $c$. The character-level LF-FontMix is defined similarly as Eq \eqref{eq:lffontmix}, while the target label is mixed as previous works \cite{mixup, yun2019cutmix}. The style mixing strategy makes complex and diverse font styles without mixing the labels. The character mixing strategy mixes images and labels at the same time, but the augmented styles can be limited. In practice, we randomly alternate style-level and character-level LF-FontMix for every iteration to take the advantages of both two strategies.

\begin{table}[b]
\small
\centering
\begin{tabular}{@{}lccc@{}}
\toprule
         & Localized & Contents & Restricted \\ 
         &  style?   & encoder? & to generate \\ \midrule
SA-VAE   & \nomark          & \nomark                & unseen chars (train)\\
EMD      & \nomark          & \yesmark               & \\
AGIS-Net & \nomark          & \yesmark               & \\
FUNIT    & \nomark          & \yesmark               & \\
DM-Font  & \yesmark         & \nomark                & unseen components (refs.) \\ 
\addition{DG-Font}   & \nomark          & \yesmark               & \\\midrule
Ours     & \yesmark         & \yesmark               &                                                                                            \\ \bottomrule
\end{tabular}
\caption{\small {\bf Comparison of \ours with other methods.} We show the taxonomy of few-shot font generation by the localized style and the content encoder. Note that SA-VAE cannot generate unseen characters during the training, and DM-Font is unable to synthesis a glyph whose component is not observable in the reference glyphs.}
\label{table:comparion_methods_desc}
\end{table}
\begin{table*}[ht!]
\centering
\small
\setlength{\tabcolsep}{4pt}
\resizebox{\textwidth}{!}{
\begin{tabular}{@{}cclccccccccccc@{}}
\toprule
                  &&                   & LPIPS $\downarrow$ &  & Acc (S) $\uparrow$ & Acc (C) $\uparrow$ & Acc (Hmean) $\uparrow$ &  & FID (S) $\downarrow$ & FID (C) $\downarrow$ & FID (Hmean) $\downarrow$ && \addition{$p_\text{unseen}$}\\ \midrule
\parbox[t]{2mm}{\multirow{7}{*}{\rotatebox[origin=c]{90}{Seen chars}}}   && SA-VAE (IJCAI'18) & 0.310 &  & 0.2       & 41.0        & 0.3         &  & 231.8     & 66.7        & 103.6   && \addition{0.01}    \\
                  && EMD (CVPR'18)     & 0.248 &  & 11.9      & 63.7        & 20.1        &  & 148.1     & 25.7        & 43.8   && \addition{0.19}      \\
                  && AGIS-Net (TOG'19) & 0.182 &  & 34.0      & \textbf{99.8}  & 50.7     &  & 79.8      & 4.0         & 7.7    && \addition{0.41}        \\
                  && FUNIT (ICCV'19)   & 0.217 &  & 39.0      & 97.1        & 55.7        &  & 58.5      & 3.6         & 6.8    && \addition{0.47}     \\
                  && DM-Font (ECCV'20) & 0.275 &  & 10.2      & 72.4        & 17.9        &  & 151.8     & 8.0         & 15.2   && \addition{0.25}     \\ %
                  
                  &&\addition{DG-Font (CVPR'21)} & \addition{0.189} &  & \addition{46.9}      & \addition{98.8}        & \addition{63.6}        &  & \addition{54.0}     & \addition{2.6}         & \addition{5.0}  && \addition{0.57}        \\ %
                  
                  && \ours (proposed)              & \textbf{0.169} &  & \textbf{75.6}      & 96.6       & \textbf{84.8}        &  & \textbf{40.4}      & \textbf{2.6}         & \textbf{4.9}     && \addition{\textbf{0.83}}    \\ \midrule
\parbox[t]{2mm}{\multirow{6}{*}{\rotatebox[origin=c]{90}{Unseen chars}}} && EMD (CVPR'18)     & 0.250 &  & 11.6      & 64.0        & 19.7        &  & 151.7     & 41.4        & 65.0    && \addition{0.19}     \\
                  && AGIS-Net (TOG'19) & 0.189 &  & 33.3      & \textbf{99.7}        & 49.9        &  & 85.4      & 10.0        & 18.0       && \addition{0.41} \\
                  && FUNIT (ICCV'19)   & 0.216 &  & 38.0      & 96.8        & 54.5        &  & 63.2      & 12.3        & 20.6        && \addition{0.46} \\
                  && DM-Font (ECCV'20) & 0.284 &  & 11.1      & 53.0        & 18.4        &  & 153.4     & 26.5        & 45.2        && \addition{0.26} \\ %
                  &&\addition{DG-Font (CVPR'21)} & \addition{0.188} &  & \addition{46.4}      & \addition{98.7}        & \addition{63.1}        &  & \addition{57.8}     & \addition{9.0}         & \addition{15.5}     && \addition{0.57}    \\ %
                  && \ours (proposed)              & \textbf{0.169} &  & \textbf{72.8}      & 97.1        & \textbf{83.2}        &  & \textbf{44.5}      & \textbf{8.7}         & \textbf{14.6} && \addition{\textbf{0.82}}       \\ \bottomrule
\end{tabular}
}
\caption{\small {\bf Performance comparison on few-shot font generation scenario.} Six few-shot font generation methods are compared with eight reference glyphs. LPIPS shows a perceptual similarity between the ground truth and the generated glyphs. We also report accuracy and FID measured by style-aware (S) and content-aware (C) classifiers. The harmonic mean (Hmean) of style- and content-aware metrics shows the overall visual quality of the generated glyphs.
\addition{$p_\text{unseen}$ indicates the ratio of the generated images correctly distinguished to unseen styles by the style classifier.}
All numbers are average of $50$ runs with different reference glyphs.}
\label{table:main_fewshot}
\end{table*}
\begin{figure*}[t]
    \centering
    \includegraphics[width=\linewidth]{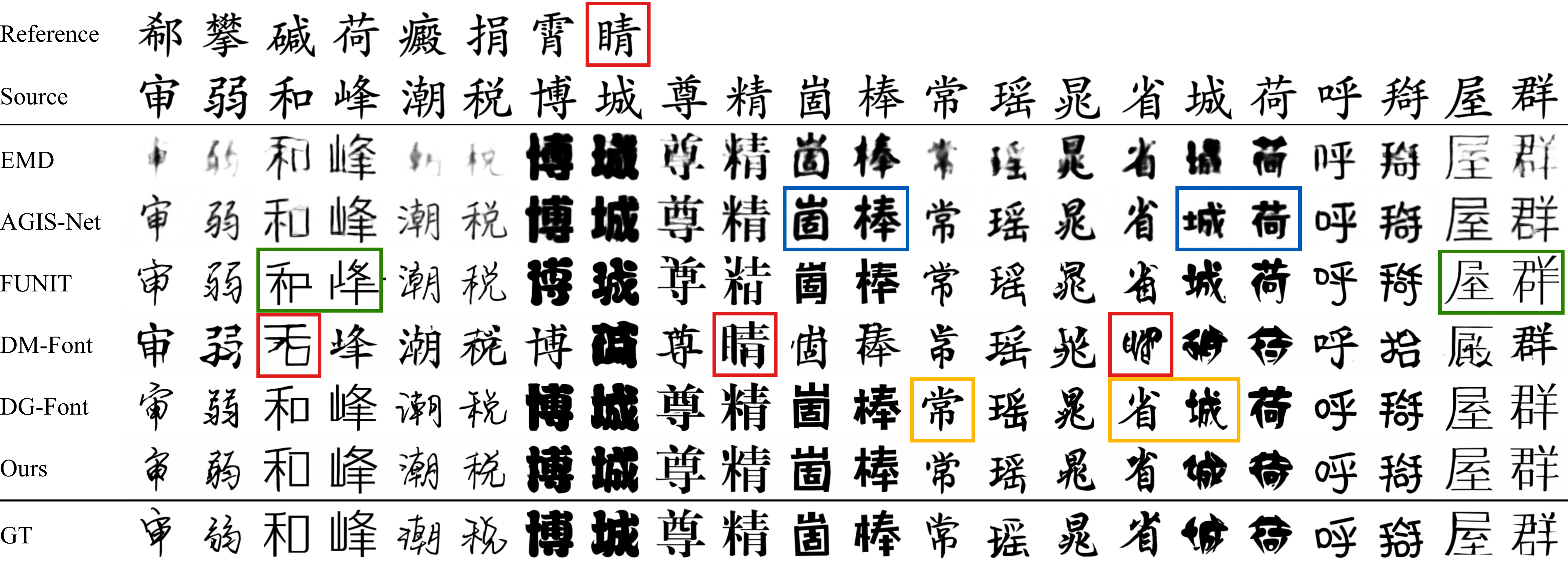}%
    \caption{\small {\bf Generated samples.} We show characters in the reference set (refer to the character only, not style), source images, generated samples of \ours and five comparison methods, and the target glyphs (see GT). The reference images for each style are shown in Appendix. We also highlight samples that show the apparent limitation of each method using colored boxes. Each color denotes the different failure cases discussed in \S~\ref{subsec:main_results}.}%
    \label{fig:main_few}
\end{figure*}

\section{Experiments}
\label{sec:experiment}
This section compares the results of \ours and previous methods for Chinese few-shot font generation. We first introduce the datasets and evaluation metrics (\S\ref{subsec:datasets}), and comparison methods (\S\ref{subsec:comparison_methods}). We compare the results of \ours and previous methods for Chinese few-shot font generation in \S\ref{subsec:main_results}. Extensive analysis on each module (\S\ref{subsec:module_study}) showed that our design choice successfully dealt with the few-shot font generation task, \eg, localized style representation (\S\ref{subsubsec:localized_style}), content encoder (\S\ref{subsubsec:content_encoder}), utilizing weakly-supervised component labels to learn complex font information (\S\ref{subsubsec:weaklysupervisedcomponentlabels}), factorization modules (\S\ref{subsubsec:factorizationmodules}), the effectiveness of style representations in extreme cases (\S\ref{subsubsec:stylerepresentationanalyses}), and two-phase training strategy (\S\ref{subsubsec:phase_abl}). We also provide experimental results with various sizes of reference images, \eg, from 1 to 256 (\S\ref{subsec:refsizestudy}) and comparisons in Korean generation tasks (\S\ref{subsec:more-lang}). Finally, in \S\ref{subsec:psuedo-character}, we present a new prediction strategy for \ours when reference images are unlabeled, \eg, the character labels are not accessible during test-time. In addition, we show the effectiveness of \ours in terms of data augmentation in \S\ref{subsec:lffontmix_results}.

\subsection{Datasets and evaluation metrics}
\label{subsec:datasets}
We collected public $482$ Chinese fonts from the Web. The dataset has a total of $19,514$ characters (each font has a varying number of characters and it is $6,654$ characters on average), which can be decomposed by $371$ components. We sample $467$ fonts corresponding to $19,234$ characters for training, and the remaining unseen $15$ fonts were used for the evaluation. The models are separately evaluated with $2,615$ {\it seen characters} and $280$ {\it unseen characters} to measure the generalizability of the unseen characters.

We evaluated the visual quality of the generated glyphs using various metrics. To measure the accuracy of the generated glyphs matching their ground truths, LPIPS~\cite{zhang2018_cvpr_lpips} with ImageNet pre-trained VGG-16 was used. LPIPS is popularly used to assess the similarity between two images by considering the perceptual similarity.

We further assessed the visual quality of the generated glyphs in two aspects: content-preserving and style-adaptation \cite{cha2020dmfont}. We trained two classifiers to distinguish the style or content labels of the test dataset. Note that we trained the evaluators independently from our generation models, and the character and font labels for the evaluation did not overlap with the training labels. ResNet-50~\cite{he2016_cvpr_resnet} was employed for the backbone architecture. Compared to photorealistic images, glyph images are highly sensitive to local damage or distinctive local component-wise information. We developed evaluation classifiers by employing CutMix augmentation~\cite{yun2019cutmix}, which leads to a model that can learn localizable and robust features~\cite{chun2019icmlw} using the AdamP optimizer~\cite{heo2020adamp}. We set a CutMix probability and the CutMix beta to 0.5 and 0.5, respectively. The batch size, learning rate, and number of epochs were set to 64, 0.0002, and 20, respectively. We report the accuracies of the generated glyphs using the {\it style-aware} and {\it content-aware} models, respectively. We also used each classifier as a feature extractor and computed the Frechét inception distance (FID)~\cite{heusel2017_nips_ttur_fid}. In the experiments, we denote metrics computed by content and style classifiers as {\it content-aware} and {\it style-aware}, respectively.

Finally, we employ a new evaluation metric that measures how the generated images can reflect a novel font style, while LPIPS and classifier-based metrics only can measure the property indirectly. We use the trained style classifier where the style classifier predicts 482 font styles, including 15 unseen test styles. After we generate images by using 15 novel font styles, we predict font styles of generated images using the trained style classifier. We report $p_\text{unseen}$ the ratio of the images predicted as unseen styles not seen styles. If the style classifier is perfect, $p_\text{unseen}$ denotes the degree of training style overfitting of the given font generation method, \eg, a lower $p_\text{unseen}$ denotes that a model is overfitted to training styles.

\subsection{Comparison methods}
\label{subsec:comparison_methods}

We compared our model with six state-of-the-art few-shot font generation methods. In this study, we did not compare our method with many-shot font generation methods, such as SC-Font \cite{jiang2019_aaai_scfont}, ChiroGAN \cite{gao2020_aaai_chirogan}, CalliGAN \cite{wu2020calligan} and RD-GAN \cite{huang2020_eccv_rdgan}, because they need a lot of reference characters and a finetuning procedure for generating each font style (\eg, SC-Font needs 755 references) and they are not able to handle unseen font styles. Our goal is to generate font libraries without an additional optimization procedure using very few references (\eg, 8 in our experiments). To understand the similarity or dissimilarity between methods, we categorize them by whether or not they explicitly model style representations or content representations, as shown in Table~\ref{table:comparion_methods_desc}.

SA-VAE~\cite{sun2018_ijcai_savae} extracts a universal style feature and utilizes a content code from the character classifier instead of the content encoder. This method cannot synthesize characters that are unseen during training.

EMD~\cite{zhang2018_cvpr_emd}, AGIS-Net~\cite{gao2019agisnet}, FUNIT~\cite{liu2019funit}, and DG-Font~\cite{xie2021dgfont} employ the content encoder, but their style representation is universal for the given style. For FUNIT, we use the modified FUNIT for the font task, as in \cite{cha2020dmfont,cha2020dmfontw}. We empirically show that this universal style representation strategy fails to capture diverse styles, even incorporating specialized modifications (\eg, the local texture discriminator) and the local texture refinement loss for AGIS-Net.

DM-Font~\cite{cha2020dmfont} is the most direct competitor to \ours. Both DM-Font and \ours utilize component-wise style features to capture local details. However, DM-Font is restricted to generating a glyph whose component is not in the reference set because it uses the learned codebook for each component instead of the content encoder. Because DM-Font generates neither Chinese characters nor glyphs with unseen components, we use the source style to extract local features for substituting the component-wise features for the unseen component.

As the original DM-Font cannot generate Chinese characters, we modified the structure of DM-Font in our Chinese few-shot generation experiments. Because Chinese characters are not decomposed into the same number of components, we modified the multi-head structure in DM-Font to a component-conditioned structure similar to that in \ours and used the averaged component-wise style features as an input to the decoder. We also changed its attention blocks to CBAM --- and eliminated the hourglass blocks in the decoder to stabilize the training. For the Korean few-shot generation experiments, we used the official DM-Font model and trained weight.

\subsection{Experimental results}
\label{subsec:main_results}
\myparagraph{Quantitative evaluation.} We evaluated the visual quality of the generated images using seven models with eight reference glyphs per style. To avoid randomness by the reference selection, we repeated the experiments $50$ times with different reference characters. A font generation method is required to satisfy two contradictory task objectives: it should preserve content and stylize well. As an extreme failure case, it performs an identity mapping, which shows the perfect content preserving score, but it will show a zero style transfer score. Hence, we report the harmonic mean of the content and style scores to probe whether a method can satisfy both objectives well. Table~\ref{table:main_fewshot} shows that our method outperforms previous state-of-the-art methods with significant gaps, for example, 20.1pp higher harmonic mean accuracy than DG-Font, and 0.3 lower harmonic mean FID than DG-Font for the unseen characters. Our method outperforms other methods in style-aware benchmarks, whereas content-aware benchmarks are not significantly damaged. For example, FUNIT, AGIS-Net, and DG-Font show comparable performance in content-aware benchmarks to \ours, but they show far lower performances than \ours in style-aware benchmarks. In other words, FUNIT, AGIS-Net, and DG-Font focus only on content preservation, while failing to achieve good stylization. We add discussions of lower content accuracies by \ours in \S\ref{sec:discussion}. Additionally, \ours shows 82\% unseen prediction ratio, which is much higher than DG-Font (57\%), FUNIT (46\%), and AGIS-Net (41\%). The results support that our localized style representation approach has benefits in learning complex local styles, achieving generalizability to novel styles, while other methods tend to memorize the styles in the training set.

\myparagraph{Qualitative evaluation.} We also qualitatively compared the generated samples using the methods in Figure~\ref{fig:main_few}. We observe that AGIS-Net often drops local details, such as serif-ness, and varying thickness (blue boxes). The green boxes show that FUNIT overly relies on the structure of the source images. Thus, FUNIT tends to destroy the local structures in the generated glyphs when the source and the overall structure of the target glyphs differ significantly. DG-Font produces good-looking results in general, but it sometimes ignores the reference style and leaves the source style as it is (yellow boxes). We argue that the universal style representation strategy of AGIS-Net, FUNIT, and DG-Font causes these problems. We further provide an extensive analysis of style representations in the latter section.

DM-Font frequently fails to generate the correct characters. For example, in the red boxes of Figure \ref{fig:main_few}, DM-Font often generates a glyph whose relative component locations are muddled. Another example is in the yellow boxes; DM-Font generates glyphs with the wrong component, observable in the references. We conjecture that the absence of the content encoder causes DM-Font to suffer from the complex structures of glyphs as we observed in \S\ref{subsubsec:content_encoder}.

Compared to others, \ours generates the most plausible results that preserve the local details of each component and the global structure of characters of target styles.

\begin{table}[ht!]
\small
\centering
\setlength{\tabcolsep}{4pt}
\revision{
\resizebox{\columnwidth}{!} {
\begin{tabular}{@{}lccccccc@{}}
\toprule
&& \multicolumn{3}{c}{Accuracies $\uparrow$} & \multicolumn{3}{c}{FIDs $\downarrow$} \\ 
Style representation $f_s$ & LPIPS $\downarrow$      & S & C & H & S & C & H  \\ \midrule
AGIS-Net                 & 0.189     & 33.3                  & \textbf{99.7}        & 49.9     & 85.4 & 10.0 & 18.0   \\
FUNIT                    & 0.216      & 38.0                  & 96.8                 & 54.5     & 63.2 & 12.3 & 20.6   \\ \midrule
Universal without $E_{s,u}$  & 0.197  & 33.6                  & 97.2         & 49.9            & 92.9 & 10.8 & 19.4   \\
Universal with $E_{s,u}$     & 0.187  & 52.8                  & 95.9         & 68.1            & 74.1 & 9.3 & 16.5    \\ \midrule
Localized with $E_{s,u}$     & 0.169 & \textbf{72.8}         & 97.1          & \textbf{83.2}   & \textbf{44.5} & \textbf{8.7} & \textbf{14.6}    \\ \bottomrule
\end{tabular}}
}
\caption{\small {\bf Impact of localized style representation.} 
The universal style without the component-wise style encoder $E_{s,u}$ is defined for each style. The universal style with $E_{s,u}$ is computed by the average of the reference component-wise styles. 
}
\label{table:abl_style}
\end{table}

\begin{table}[ht!]
\small
\centering
\setlength{\tabcolsep}{5pt}
\revision{
\resizebox{\columnwidth}{!} {
\begin{tabular}{@{}lccccccc@{}}
\toprule
&& \multicolumn{3}{c}{Accuracies $\uparrow$} & \multicolumn{3}{c}{FIDs $\downarrow$} \\
& LPIPS $\downarrow$ & S & C & H & S & C & H \\ \midrule
& \multicolumn{7}{c}{Few-shot} \\ \midrule
DM-Font & 0.284 & 11.1 & 53.0 & 18.4 & 152.5 & 26.3 & 44.8  \\
\ours without $E_c$ & 0.255 & 36.3 & 15.4 & 21.7 & 100.8 & 28.3 & 44.2  \\
\ours & 0.169 & \textbf{72.8} & \textbf{97.1} & \textbf{83.2} & \textbf{44.5} & \textbf{8.7} & \textbf{14.6} \\ \midrule
& \multicolumn{7}{c}{Many-shot} \\ \midrule
DM-Font & 0.254 & 51.8 & 15.0 & 23.2 & 76.3 & 25.3 & 38.0 \\
\ours without $E_c$ & 0.262 & 37.8 & 5.1 & 8.9 & 97.5 & 30.3 & 46.3 \\
\ours &  0.165 & \textbf{74.7} & \textbf{96.5} & \textbf{84.2} & \textbf{41.4} & \textbf{8.6} &\textbf{14.3} \\
\bottomrule
\end{tabular}
}
}
\caption{\small {\bf Impact of content representation.} We evaluate DM-Font, \ours without content encoder $E_c$, and \ours, in the few- (8 references) and many-shot (256 references) scenarios.
}
\label{table:abl_content} 
\end{table}
\begin{figure}[ht!]
    \centering
    \includegraphics[width=.7\linewidth]{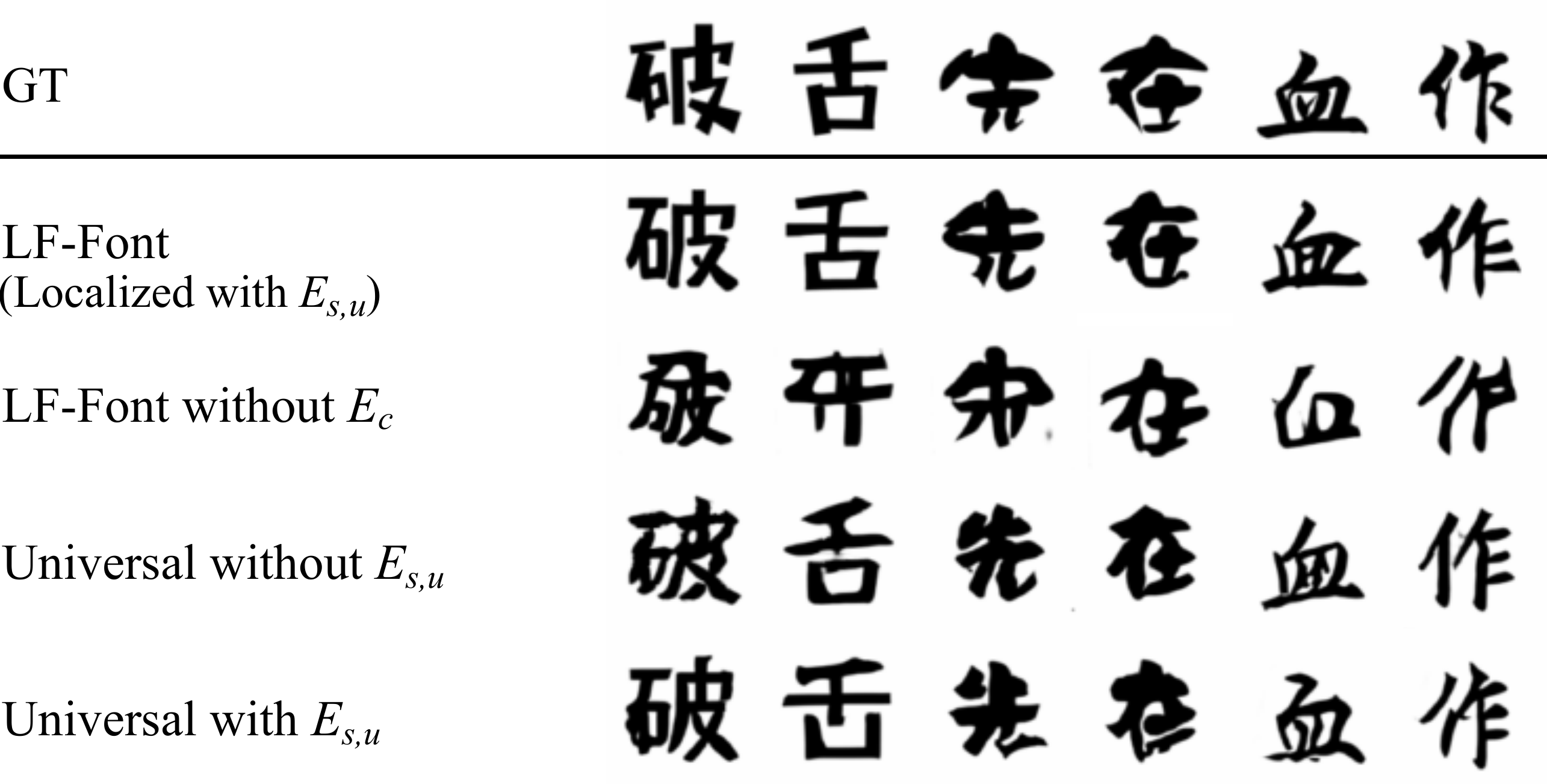}%
    \caption{\small {\bf Visual samples of style and content module analysis.} The visual samples in Table~\ref{table:abl_content} and Table~\ref{table:abl_style} are shown.}%
    \label{fig:example_ablation}
\end{figure}

\subsection{Module and parameter analyses}
\label{subsec:module_study}

\subsubsection{Localized vs. universal style representation}
\label{subsubsec:localized_style}
We compare two universal style encoding strategies to our localized style encoding strategy. First, we train a universal style encoder that extracts a universal style from the references. EMD, AGIS-Net, and FUNIT have employed this scheme. We also develop an alternative universal style encoding strategy with a component-wise style encoder $E_{s,u}$. This alternative encoding utilizes $E_{s,u}$ to extract component-wise features from references; however, the extracted features are directly used without considering the target character. On the other hand, our localized style encoder encodes character-wise localized style representations using $E_{s,u}$ and factorization modules.

We conducted an ablation study to investigate the effects of different style encoding strategies and summarize the results in Table~\ref{table:abl_style} (the same evaluation setting as Table~\ref{table:main_fewshot}). In Table~\ref{table:abl_style}, we observe that the universal style encoding without $E_{s,u}$ shows comparable style-aware performance (33.6\%) to AGIS-Net (33.3\%) --- or FUNIT (38.0\%). Also, the universal style representations by adding the component-wise style encoder $E_{s,u}$ is useful for increasing the style-aware metric (33.6\% $\rightarrow$ 52.8\%); and our reorganized localized style representation improves the style-aware metric (33.6\% $\rightarrow$ 72.8\%). The generated samples for each ablation are shown in Figure~\ref{fig:example_ablation}. From these results, we conclude that the proposed localized style representation enables the model to capture diverse local styles, whereas the universal style encoding fails in fine and local styles.

\subsubsection{Content encoder}
\label{subsubsec:content_encoder}
Although localized style encoding brings remarkable improvements in transferring a target style, our strategy has a drawback: it will extract the same feature for characters whose components are identical, but the locations vary. To solve this problem, we employ the content encoder $E_c$ to capture the structural information. Here, we examine various content-encoding strategies: \ours without content-encoding (generating the target glyph with localized style features alone), DM-Font ({\it persistent memory} for content encoding), and \ours. DM-Font replaces the content encoder with {\it persistent memory}, which is a learned codebook defined for each component. Note that DM-Font cannot generate unseen reference components; thus, we replace the unseen component features with the source style features. To remove unexpected effects from this source style replacement strategy, we reported many-shot (256 references) results in addition to the few-shot (8 references) results.

In Table~\ref{table:abl_content}, we observe that the content encoder notably enhances the overall performance (21.7\% $\rightarrow$ 83.2\% in few-shot harmonic mean accuracy). Because there is no content information, the style encoder of \ours without $E_c$ should encode both the style and content information of each component. However, as the style encoder is optimized for modeling local characteristics, it is limited to handling global structures (\eg, the positional relationship of components). In addition, because a combination of a component set can be mapped to diverse characters, as shown in Figure~\ref{fig:twins}, solely learning localized style features without global structures cannot reconstruct the correct character even though it can capture detailed local styles. Qualitative examples for \ours without the content encoder are shown in Figure~\ref{fig:example_ablation}.

Similar to the content encoder, the persistent memory strategy proposed by DM-Font, moderately improves the content performance (15.4\% $\rightarrow$ 53.0\%); but shows worse stylization owing to the source style replacement strategy. Furthermore, both \ours without $E_c$ and DM-Font suffer from a content performance drop in the many-shot setting. This is because their style encoders suffer from encoding the complex structures --- \eg, relative size or positions --- of the unseen styles, as shown in Figure~\ref{fig:main_few} (the yellow boxes).

\begin{table}[t!]
\small
\centering
\revision{
\resizebox{\columnwidth}{!} {
\begin{tabular}{@{}ccccccccc@{}}
\toprule
&&& \multicolumn{3}{c}{Accuracies $\uparrow$} & \multicolumn{3}{c}{FIDs $\downarrow$} \\ 
$\mathcal L_{consist}$ & $\mathcal L_{cls}$ & LPIPS $\downarrow$ & S & C & H & S & C & H \\ \midrule
\nomark  & \nomark   & 0.206 & 44.5             & 76.3          & 56.2    & 82.1 & 15.0 & 25.4\\
\yesmark & \nomark   & 0.195 & 47.2             & 88.6          & 61.6    & 77.9 & 10.7 & 18.8\\
\nomark  & \yesmark  & \textbf{0.169} & 69.3             & \textbf{97.2} & 81.1    & 49.8 & \textbf{8.6} & 14.7        \\ \midrule
\yesmark & \yesmark  & \textbf{0.169} & \textbf{72.8}    & 97.1          & \textbf{83.2} & \textbf{44.5} & 8.7 & \textbf{14.6}    \\ \bottomrule
\end{tabular}}
}
\caption{
\small {\bf Impact of objective functions.}
We report the results of the different combinations of consistency loss $L_{consist}$ and the component-classification loss $L_{cls}$.
Our design choice is the bottom row, which shows the best overall performance.}
\label{table:abl_loss}
\end{table}
\begin{table}[t!]
\small
\centering
\revision{\begin{tabular}{@{}ccccccccc@{}}
\toprule
& & \multicolumn{3}{c}{Accuracies $\uparrow$} && \multicolumn{3}{c}{FIDs $\downarrow$} \\ 
$k$             &  LPIPS $\downarrow$  & S & C & H && S & C & H        \\ \midrule
4               & 0.169        & 71.0          & 98.0              &82.3             && 46.1 & 8.8 & 14.8  \\
6               & 0.168        & 72.0          & \textbf{98.0}     &83.0             && 44.6 & \textbf{8.6} & 14.5\\
8$^\dagger$     & 0.169        & \textbf{72.8} & 97.1              & \textbf{83.2}   && \textbf{44.5} & 8.7 & 14.6  \\
10              & \textbf{0.167}        & 71.4          & 97.5              & 82.4            && 45.6 & \textbf{8.6} & \textbf{14.4 } \\ \bottomrule
\end{tabular}}
\caption{
\small {\bf Factor size study.} 
$^\dagger$ used in the remaining results.}
\label{table:abl_k}
\end{table}

\subsubsection{Weakly supervised component labels}
\label{subsubsec:weaklysupervisedcomponentlabels}
We analyzed the importance of the image-level component label as weak supervision to learn localized style representations. We utilized the component labels in two modules. First, we use the component labels for the component-wise style encoder $E_{s, u}$, whose importance has already been demonstrated in the previous section. The other module is the auxiliary component classifier $Cls$, which guides the local features extracted by $E_{s, u}$. We conducted a loss ablation study and demonstrated the effect of utilizing weak component-level supervision by $\mathcal L_{cls}$. Table~\ref{table:abl_loss} illustrates that utilizing the component supervision is a critical factor for capturing both diverse local style and global content structure; the performance gains are significant such as 47.2\% $\rightarrow$ 72.8\% for the style accuracy, 88.6\% $\rightarrow$ 97.1\% for the content accuracy, and 61.6\% $\rightarrow$ 83.2 for their harmonic mean, respectively. Here, using only $E_{s,u}$ still produces better style-aware performance (47.2\%) than AGIS-Net (33.3\%) and FUNIT (38.0\%), but the content-aware performance is degraded. We speculate that this is because of insufficient component supervision. We observe that $\mathcal L_{cls}$ has a particularly large impact on the style-aware and content-aware performance: 47.2\% $\rightarrow$ 72.8\%, 88.6\% $\rightarrow$ 97.2\%. These results demonstrate that, to improve the overall performance, employing $Cls$ following $E_{s,u}$ plays a crucial role because the classifier provides sufficient component supervision to the model.

\subsubsection{Factorization modules}
\label{subsubsec:factorizationmodules}
The factorization modules ($F_s, F_u$) takes a key role in \ours in terms of the generalizability to novel styles. In this subsection, we show the reconstruction capability of the full feature set from the given features according to the presence of constraints and factor size.

We first report the performance of the models when the factors are unconstrained, that is, $z_u$ and $z_s$ are not unique for component $u$ and style $s$ in Table~\ref{table:abl_loss}. The factors constrained by $\mathcal L_{consist}$ improve the overall performance; 69.3\% $\rightarrow$ 72.8\% for the style accuracy, 97.2\% $\rightarrow$ 97.1\% for the content accuracy, and 81.1\% $\rightarrow$ 83.2 \% for the harmonic mean. These results show that the constrained factors contribute to the performance improvements but are less effective than the presence of weak component-level supervision, $\mathcal L_{cls}$.

Table~\ref{table:abl_k} shows performances by varying the factor size $k$ from 4 to 10. We observe that a larger $k$ enhances the harmonic mean performance (82.3\% $\rightarrow$ 83.2\% when we set $k$ as 4 $\rightarrow$ 8), but the overall performance converges after $k \geq 8$. For the efficiency, we use $k=8$ in this paper.

\subsubsection{Style representation analyses}
\label{subsubsec:stylerepresentationanalyses}
\begin{figure}[t]
    \centering
    \includegraphics[width=0.8\linewidth]{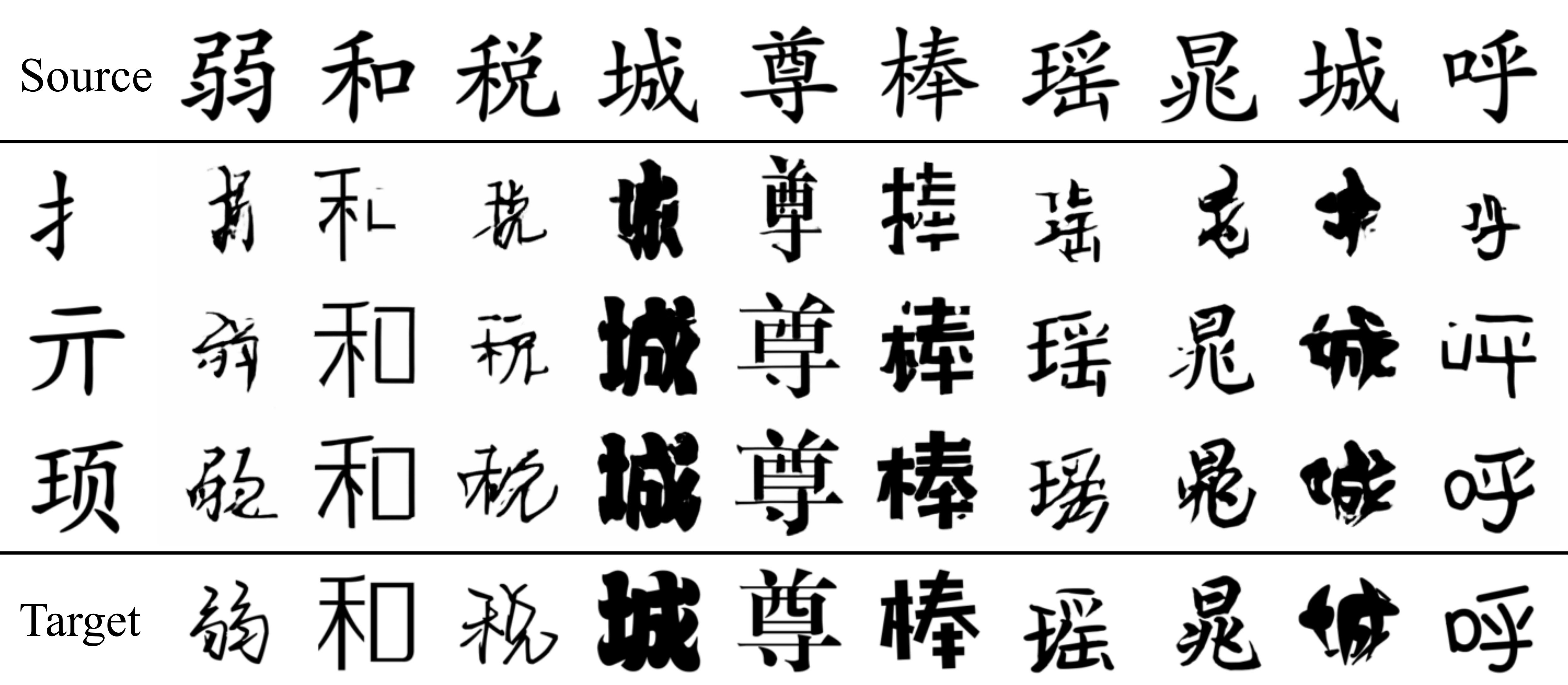}%
    \caption{\small {\bf One-shot generation results.} The reference characters and the resultant images were visualized. The top and bottom rows show the source and target images; and the leftmost column shows the single reference used to generate the images.
    }
    \label{fig:1shot}
\end{figure}

\myparagraph{One-shot generation.} 
We visualize the extreme case, the one-shot generation, shown in Figure~\ref{fig:1shot}. We observe that when the reference glyph is too simple to extract solid component-wise features (the second row in Figure~\ref{fig:1shot}), the generated images show poor visual quality. This may be an issue concerning style factors, not $E_c$, because the same content features (from $E_c$) are used to successfully generate other samples. Hence, the reference selection with rich local details is critical to high-quality generation.

\begin{figure}[t]
    \centering
    \begin{subfigure}[b]{0.48\linewidth}
        \includegraphics[width=\linewidth]{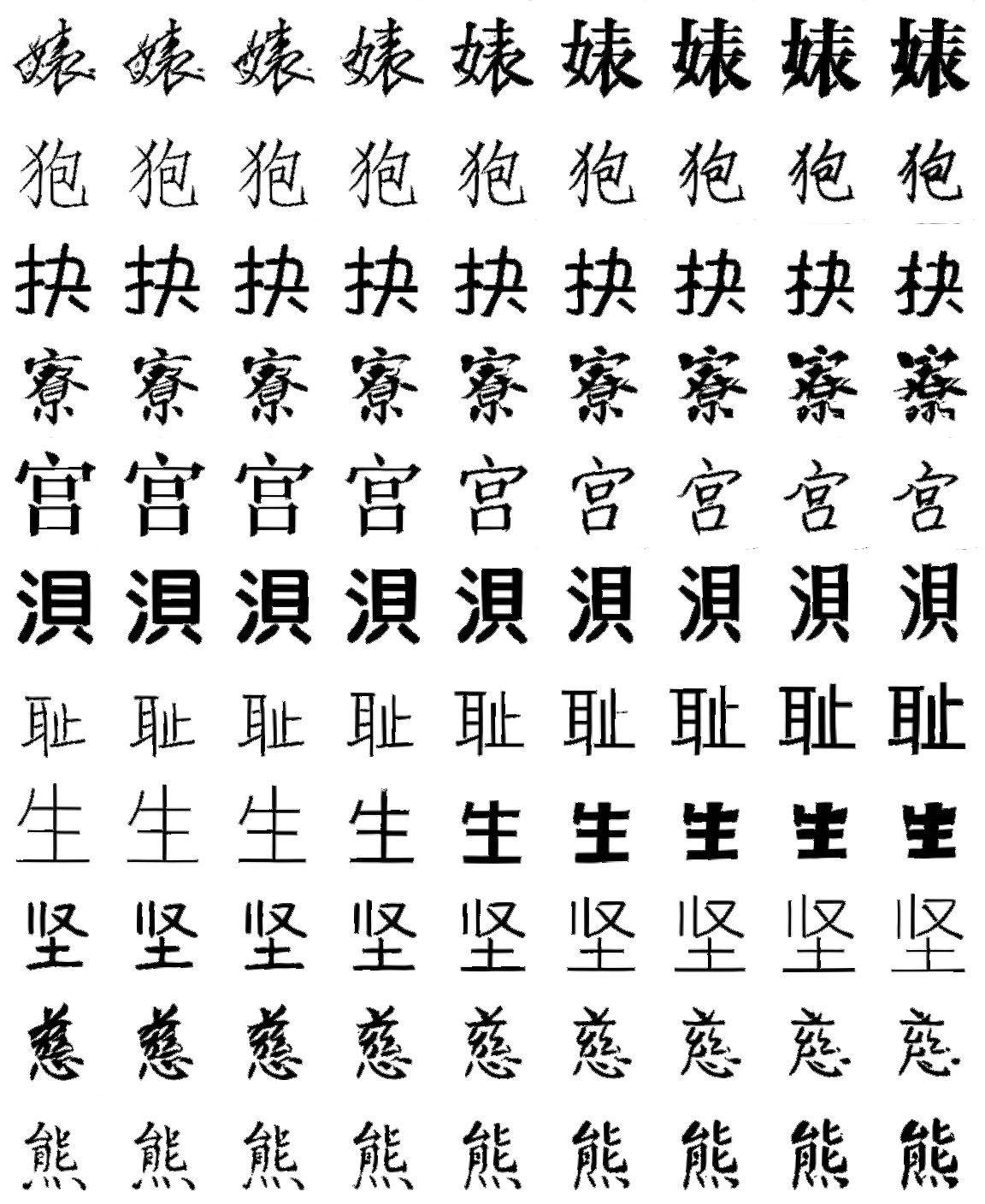}%
        \caption{\small Style interpolation}
        \label{fig:interp_b}
    \end{subfigure}
    \hfill
    \begin{subfigure}[b]{0.48\linewidth}
        \includegraphics[width=\linewidth]{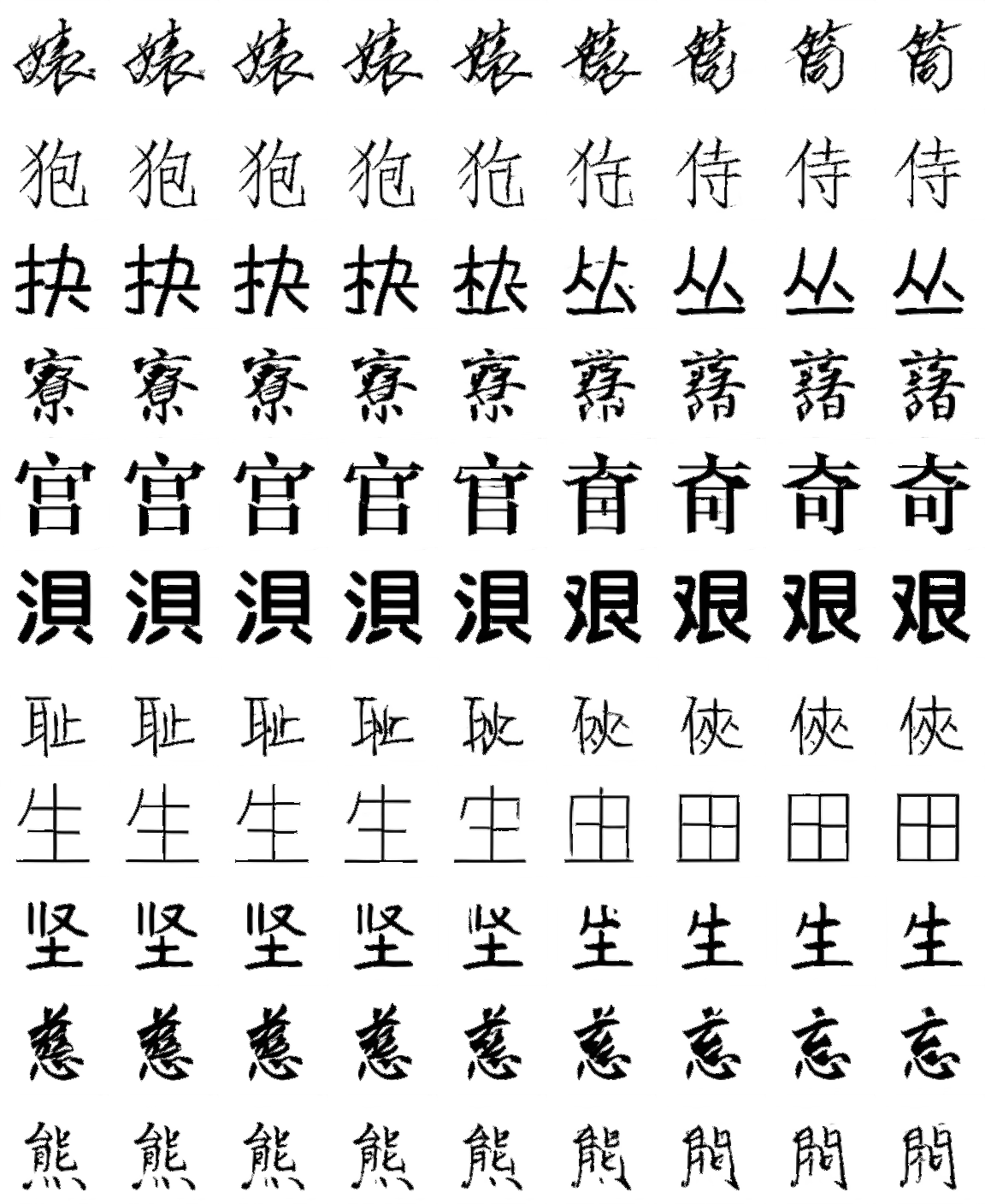}%
        \caption{\small Character interpolation}
        \label{fig:interp_c}
    \end{subfigure}
    \caption{\small {\bf Interpolation results.} The interpolated images by \ours are shown. For each figure, we mix two images from the leftmost and the rightmost images. We provide three different mixing strategies (a) mixing images in style representation only and (b) mixing images in content representations only.}%
    \label{fig:interp}
\end{figure}

\myparagraph{Style and character interpolation.}
Figure~\ref{fig:interp_b} shows the style-interpolated images; we linearly interpolate only the style factors $z_s$ extracted from the character-wise style features while remaining the component factors $z_c$ and the content representations. We interpolate images in the character-level similarly (Figure \ref{fig:interp_c}); we interpolated the component factors $z_u$ and the content representations while remaining the style factors $z_s$. The style interpolation results show that \ours provides semantically meaningful style features thus presents well-interpolated local features. A smooth transition across two different styles and content also demonstrates that \ours leads to well-disentangled content-style representations while capturing diverse font-specific characteristics and the character contents.

\begin{figure*}[ht!]
    \centering
    \includegraphics[width=.85\linewidth]{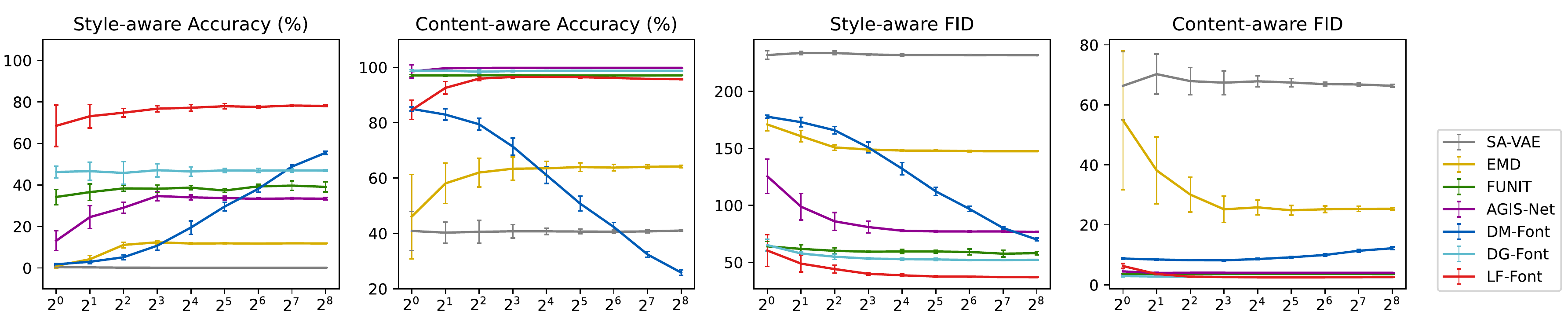}%
    \caption{
    \small{\bf Performance changes by varying size of the reference set.} We report how the performance of each model is affected by the size of the reference set $\mathcal X_r$. The style-aware performance and content-aware performance are evaluated by generating seen characters, and each recorded two metrics: accuracy (higher is better), and FID (lower is better). Each graph shows the average performance as a line, and errors as an error bar.
}
    \label{fig:numref_performance}
\end{figure*}

\begin{figure*}[ht!]
    \centering
    \includegraphics[width=.85\linewidth]{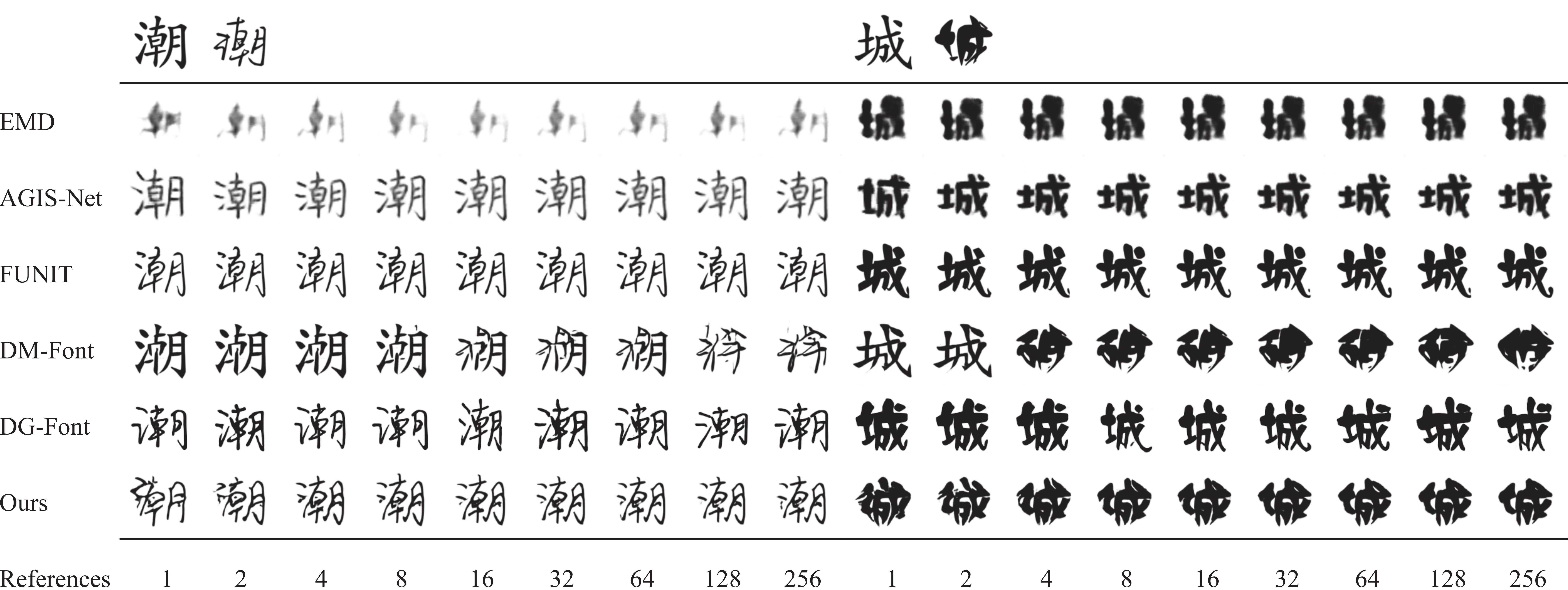}%
    \caption{
    \small {\bf Generated samples by varying reference set size.} Each row shows the samples generated by each model. The source and target glyphs are displayed in the top row.}%
    \label{fig:nref_sample}
\end{figure*}

\begin{figure*}[t]
    \centering
    \includegraphics[width=.85\linewidth]{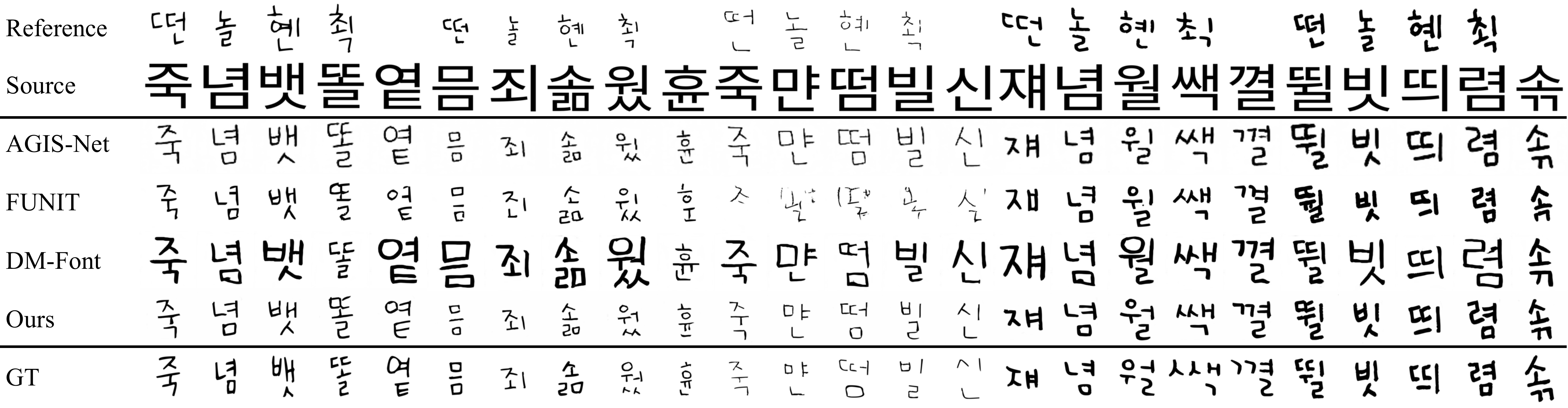}%
    \caption{
    \small {\bf Korean few-shot generation samples.} The samples generated by each model and ground truth glyphs are shown. The samples were generated with four reference images, which are shown in the top row.}%
    \label{fig:kor_result}
\end{figure*}

\begin{table}[ht!]
\small
\centering
\setlength{\tabcolsep}{4pt}
\addition{
\begin{tabular}{@{}lccccccc@{}}
\toprule
& & \multicolumn{3}{c}{Accuracies $\uparrow$} & \multicolumn{3}{c}{FIDs $\downarrow$} \\ 
Character label  & LPIPS $\downarrow$  & S & C & H & S & C & H  \\ \midrule
Prediction   & 0.171 & \textbf{74.1}                 & 94.8        & \textbf{83.2}     & \textbf{44.3} & 8.8 & 14.7   \\
Ground-truth      & 0.169 & 72.8         & \textbf{97.1}          & \textbf{83.2}   & 44.5 & \textbf{8.7} & \textbf{14.6}    \\ \bottomrule
\end{tabular}
}
\caption{\addition{
\small {\bf Comparison of different character label rules on test-time.} ``Prediction'' denotes that \ours uses the predictions by an auxiliary character classifier, while ``Ground-truth'' uses the ground-truth character labels directly.
}}
\label{table:abl_pred_char}
\end{table}
\begin{table}[ht!]
\small
\centering
\setlength{\tabcolsep}{4pt}
\addition{
\resizebox{\columnwidth}{!} {
\begin{tabular}{@{}lccccccc@{}}
\toprule
& & \multicolumn{3}{c}{Accuracies $\uparrow$} & \multicolumn{3}{c}{FIDs $\downarrow$} \\ 
Training   & LPIPS $\downarrow$ & S & C & H & S & C & H  \\ \midrule
End-to-end  & 0.208 & 36.6                 & 91.3        & 52.2     & 663.8 & 15.3 & 30.8   \\
Two-phase (proposed)     & 0.169 & \textbf{72.8}         & \textbf{97.1}          & \textbf{83.2}   & \textbf{44.5} & \textbf{8.7} & \textbf{14.6}    \\ \bottomrule
\end{tabular}
}
}
\caption{\addition{
\small {\bf Ablation of two-phase learning strategy.}
We compare the proposed two-phase learning strategy and the end-to-end learning with only a single phase.
}}
\label{table:abl_phase}
\end{table}
\subsubsection{Two-phase training vs. end-to-end training}
\label{subsubsec:phase_abl}
We show the effectiveness of the proposed two-phase training strategy compared to the end-to-end single-phase strategy. For the end-to-end training, we mix mini-batch selection policies of phase 1 and 2. Table \ref{table:abl_phase} shows the results. Our two-phase training strategy shows much more stable convergence than the end-to-end strategy. As we discussed in \S\ref{subsec:training}, our two-phase training strategy helps the factorization modules to learn factorization rules on stable features not highly varying features.

\begin{table*}[ht!]
\small
\centering
\begin{tabular}{@{}lccccccc@{}}
\toprule
                    & LPIPS $\downarrow$ & Acc (S) $\uparrow$ & Acc (C) $\uparrow$ & Acc (Hmean) $\uparrow$ & FID (S) $\downarrow$ & FID (C) $\downarrow$ & FID (Hmean) $\downarrow$ \\ \midrule
AGIS-Net (TOG'19)   & 0.188        & 3.9 & 97.5 & 7.5 & 108.1 & 7.8 & 14.5 \\
FUNIT (ICCV'19)   & 0.202        & 7.3 & 85.1 & 13.4 & 68.4 & 9.6 & 16.8 \\
DM-Font (ECCV'20)   & 0.266        & 3.4 & 96.3 & 6.5 & 126.3 & 19.0 & 33.0 \\
LF-Font (proposed)  & \textbf{0.145} & \textbf{41.6} & \textbf{98.4} & \textbf{58.5} & \textbf{47.2} & \textbf{4.9} & \textbf{8.9} \\ \bottomrule
\end{tabular}
\caption{
\small {\bf Performance comparison on few-shot Korean font generation scenario.} We report LPIPS, FID and accuracy measures for AGIS-Net, DM-Font and LF-Font. All numbers are average of 10 runs with different reference glyphs.}
\label{table:app_kor}
\end{table*}

\subsection{Reference size study}
\label{subsec:refsizestudy}
Figure \ref{fig:numref_performance} shows the performance of few-shot methods by varying the number of the references from 1 to 256. In the style-aware metrics, \ours performs remarkably better than all other methods. Surprisingly, \ours using one reference outperforms others using many references (\eg 256.) in style-aware metrics. Despite the impressive style-aware performance, \ours shows less stable content-aware performances than FUNIT, AGIS-Net, and DG-Font in the low reference regime. However, we observe that the samples generated by FUNIT, AGIS-Net, and DG-Font with very few references are rarely stylized, but only maintain the source shape in Figure~\ref{fig:nref_sample}. As shown in the one-shot generation results (Figure \ref{fig:1shot}), when the number of references is extremely small, \ours can be sensitive to the reference sampling, that is, more complex references provide rich local representations, --- thus improving generation performance.

\subsection{Extension to other languages}
\label{subsec:more-lang}
We report the Korean few-shot generation results by \ours, AGIS-Net~\cite{gao2019agisnet}, and DM-Font~\cite{cha2020dmfont} with four reference glyphs in Table~\ref{table:app_kor}. We used the same training and test datasets used by Cha~\etal~\cite{cha2020dmfont}, as well as the evaluation classifiers. Notably, Cha~\etal used a larger reference size (28 references) and employed a special sampling strategy in which the sampled reference set covers the complete component labels. Following our main experimental protocol, we report the averages of LPIPS, FID, and accuracies of ten different runs with different reference selections to reduce the randomness to the reference selection. In the table, we observe that LF-Font outperforms DM-Font and AGIS-Net in terms of overall metrics, particularly in {\it style-aware} metrics. The visual examples are illustrated in Figure~\ref{fig:kor_result}.

\subsection{Generation with pseudo-character labels}
\label{subsec:psuedo-character}
\ours requires explicit character labels (or component labels) even in test-time. In this subsection, we compare the pseudo-character label-based prediction strategy (discussed in \S\ref{subsec:generation}) to the ground-truth character labels. We evaluate our model in two different scenarios. First, we test the models in the \textbf{in-domain transfer scenario}, \ie, training a model on Chinese images and evaluating a model on Chinese images as same as Table \ref{fig:main_few}. Second, the models are evaluated on \textbf{zero-shot cross-lingual scenario}; we train a model on Chinese glpyh images and evaluating the model by generating Korean images. 

\myparagraph{In-domain transfer scenario.}
Table \ref{table:abl_pred_char} shows the comparison of the ground-truth character label rule and prediction-based character-pseudo label rule in Chinese-to-Chinese generation scenario. Note that we use the same \ours model trained on Chinese script for each character label rule. In the table, the prediction-based solution shows slightly lower content performances due to the inherent classification errors by our auxiliary character classifier. Interestingly, we observe that the prediction-based solution shows slightly better style performances than the GT solution, thus two methods show similar harmonic mean performances. It implies that the correct component condition is not necessary for better stylization, but it is required for better content-preserving. We add related discussions in \S\ref{sec:discussion}.

\myparagraph{Zero-shot cross-lingual scenario.}
We slightly modify the generation procedure of \ours to handle unseen language systems by omitting component conditions from the component-conditioned encoder. This enables the representation of \ours to have universal style. Table \ref{table:zeroshot_results} shows that the performances on zero-shot cross-lingual (Chinese-to-Korean) generation tasks are significantly improved by our prediction-based strategy. Here, we only report accuracies because FID and LPIPS need ground-truth target characters of the given style but our test fonts do not have the target Korean characters. Note that since Korean and Chinese do not share their components, the vanilla \ours cannot utilize the power of localized style representations in this scenario. The Chinese-to-Korean generation results are also aligned with Table \ref{table:abl_style}; our localized style representation is the key of the high-quality generation performances.

\begin{table}[ht!]
\small
\centering
\setlength{\tabcolsep}{4pt}
\addition{
\begin{tabular}{@{}lccccc@{}}
\toprule
&  Vanilla & CutMix & Ours (S) & Ours (C) & Ours (S,C) \\ \midrule
Accuracy &	45.4 &	51.6 & 62.4 & 63.9 & 66.8 \\ \bottomrule
\end{tabular}
}
\caption{\addition{
\small {\bf Comparison of different font augmentations.} Vanilla, CutMix and LF-FontMix (Ours) are shown.
}}
\label{table:lffontmix}
\end{table}

\subsection{Font generation as data augmentation}
\label{subsec:lffontmix_results}
LF-Font shows plausible style- or character-interpolated images for the given two glyph images (Figure \ref{fig:interp}), showing its potential as the effective augmentation policy, LF-FontMix. We compare our LF-FontMix (\S\ref{subsec:lffontmix}) to the vanilla strategy (without any augmentation) and CutMix \cite{yun2019cutmix}. We train character classifiers that predict 6,166 distinct characters using 5 images per character, \ie, when styles are diverse and each style has very few images. The classifiers are optimized by the AdamP optimizer \cite{heo2020adamp} for 90 epochs with the initial learning rate 0.0002 decayed by the cosine annealing scheduler. The batch size is 256. For all mix-based strategies, the half of mini-batch images are mixed while the remaining half images are used as the original. The mix combination ratio $\lambda$ is sampled from Beta\,(0.5, 0.5).

We investigate three variations of LF-FontMix; style-mix, character-mix and style-character-mix (\S\ref{subsec:lffontmix}). We compare the vanilla, CutMix, and LF-FontMix strategies in Table \ref{table:lffontmix}. LF-FontMix remarkably improves character recognition performances compared to the vanilla (45.4\% $\rightarrow$ 66.8\%) and CutMix (51.6\% $\rightarrow$ 66.8\%). We also confirm that solely adopting the style-mix or the character-mix enhances the performances. Interestingly, although style-mix does not mix labels as CutMix or character-mix, it achieves higher accuracy (62.4\%) than the vanilla and CutMix by augmenting diverse font styles, while LF-FontMix (C) achieves better accuracy by mixing images and labels at the same time (63.9\%). Our LF-FontMix (S,C) takes the advantages of style-mix and character-mix, achieving the best accuracy (66.8\%).

\section{Discussion and Limitations}
\label{sec:discussion}

In this section, we discuss the limitations of \ours and the possible future research directions.

\myparagraph{Low content accuracies and failure cases on the low-shot scenario.}
In Table \ref{table:main_fewshot}, \ours shows lower content accuracies than other methods, \eg, LF-Font shows 96.6\% unseen character accuracy while AGIS-Net shows 99.7\% accuracy. There are two aspects why \ours shows lower content classification accuracies than others. First, \ours focused on learning various local styles. As shown in $p_\text{unseen}$ in Table \ref{table:main_fewshot}, \ours is capable of being generalized to novel styles while other methods heavily rely on the training styles and focus on content preserving. The qualitative results in Figure \ref{fig:main_few} show that the comparison methods overly rely on the structure of the source image. In other words, there exists the trade-off between content preserving and stylization in font generation tasks, \eg, if a model generates source images without any stylization, the content accuracies will be always 100\%. Hence it is important to simultaneously consider two different metrics in terms of content score and style score to evaluate font generation methods. Our \ours shows the best harmonic mean accuracy on both seen and unseen characters.

\begin{figure}[t]
    \centering
    \includegraphics[width=0.65\linewidth]{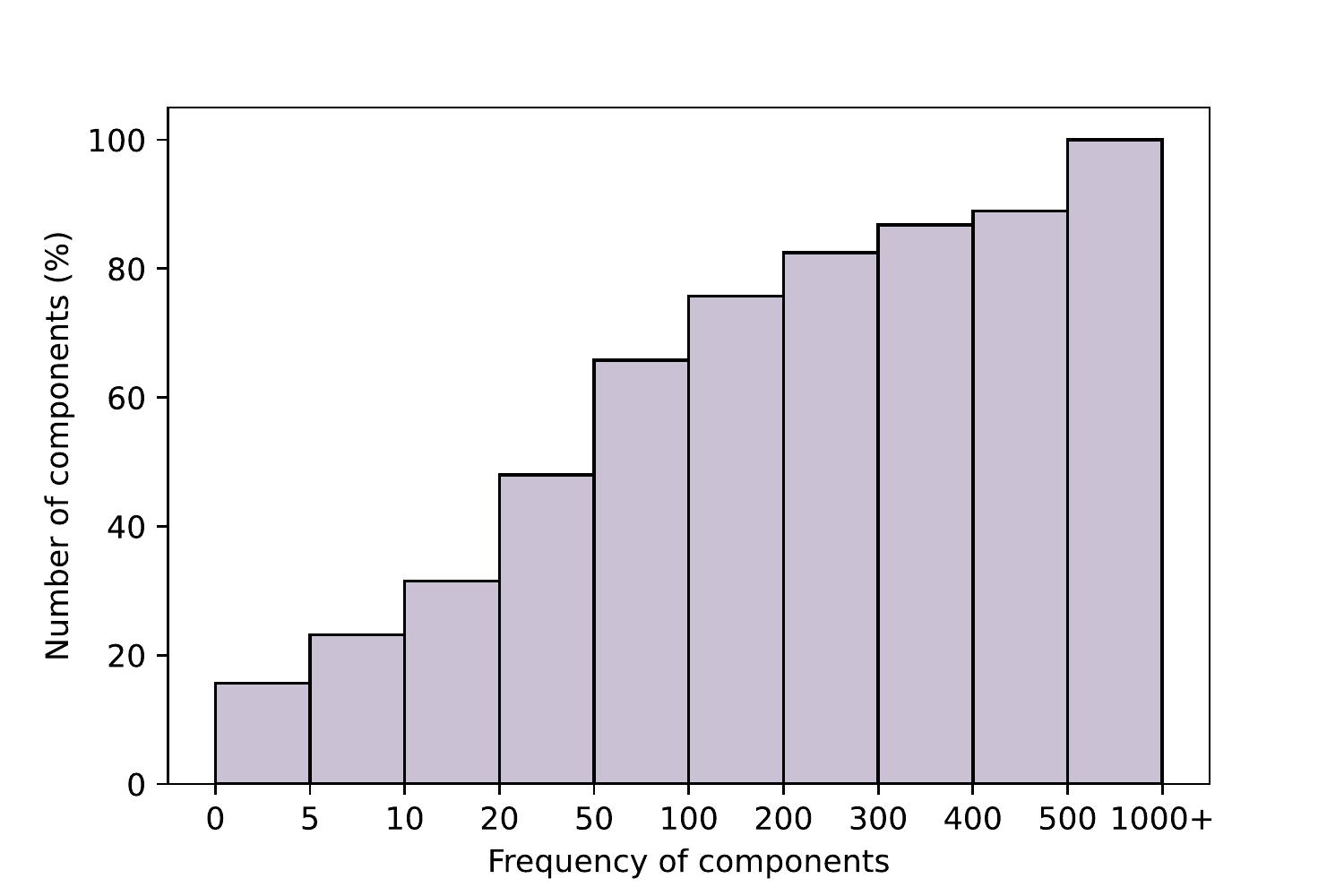}%
    \caption{\small {\bf Component frequency in 19,234 Chinese characters.} Over 50\% of components only appear in less than 50 characters.}%
    \label{fig:comp_freq}
\end{figure}

\begin{figure}[t]
    \centering
    \includegraphics[width=.7\linewidth]{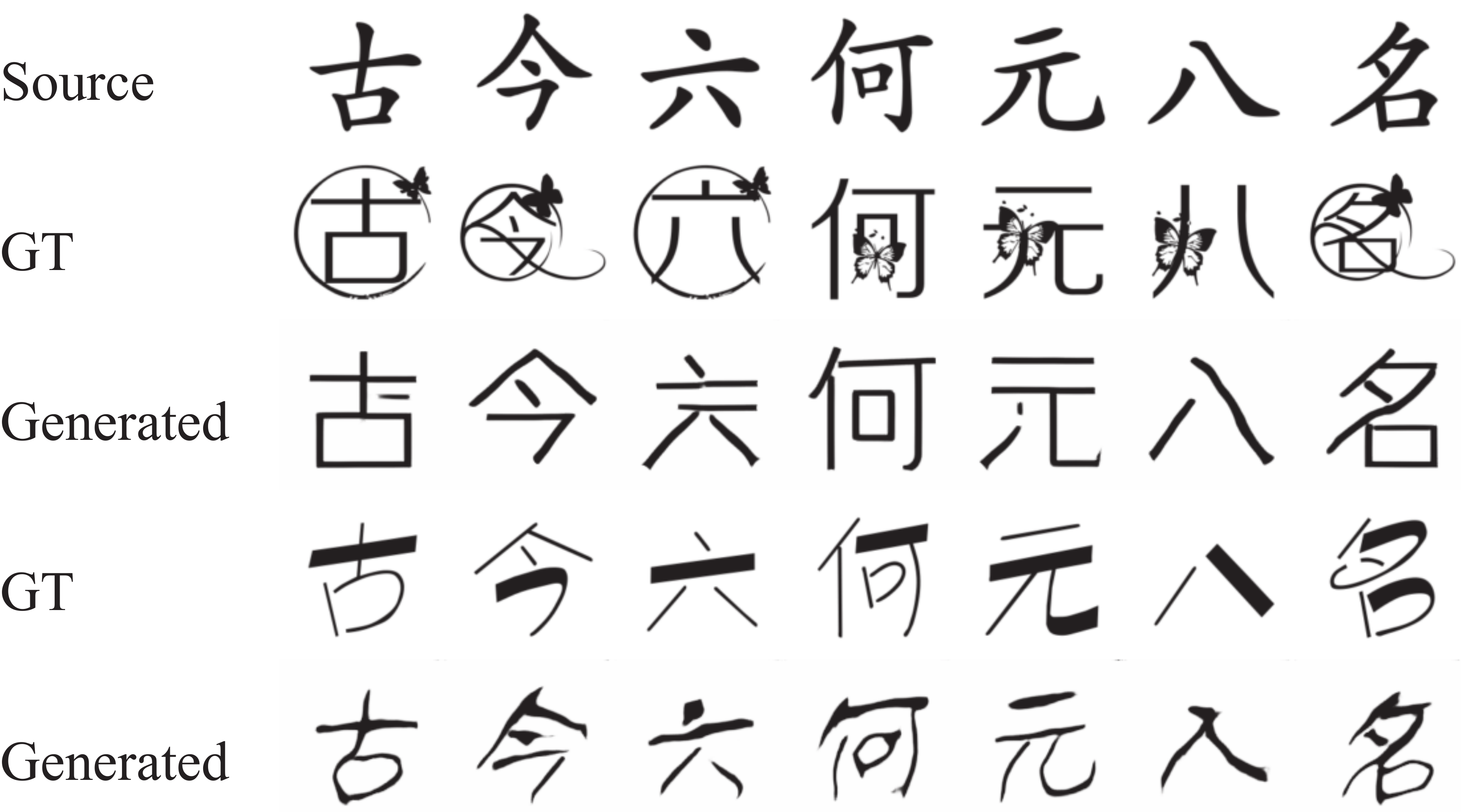}%
    \caption{\small {\bf Generation results of \ours on very novel fonts.}}%
    \label{fig:novel_fonts}
\end{figure}

Nonetheless, we also observe that our method can fail to generate very novel styles with very few references. We report two cases, 1) in Figure \ref{fig:1shot}, we observe that if the reference components are too simple, \ours fails to preserve the complex structure of the source characters. 2) in Figure \ref{fig:novel_fonts}, \ours fails to generate images with very novel font style, such as a font with a novel decoration (circles and butterflies in the top row) or a font with non-uniform outlines (in the bottom row). We presume that it is because the component frequency skewness of Chinese script hinders the training of our component-wise encoder; Chinese script shows long-tailed components (Figure~\ref{fig:comp_freq}).

\begin{figure}[t]
    \centering
    \includegraphics[width=.7\linewidth]{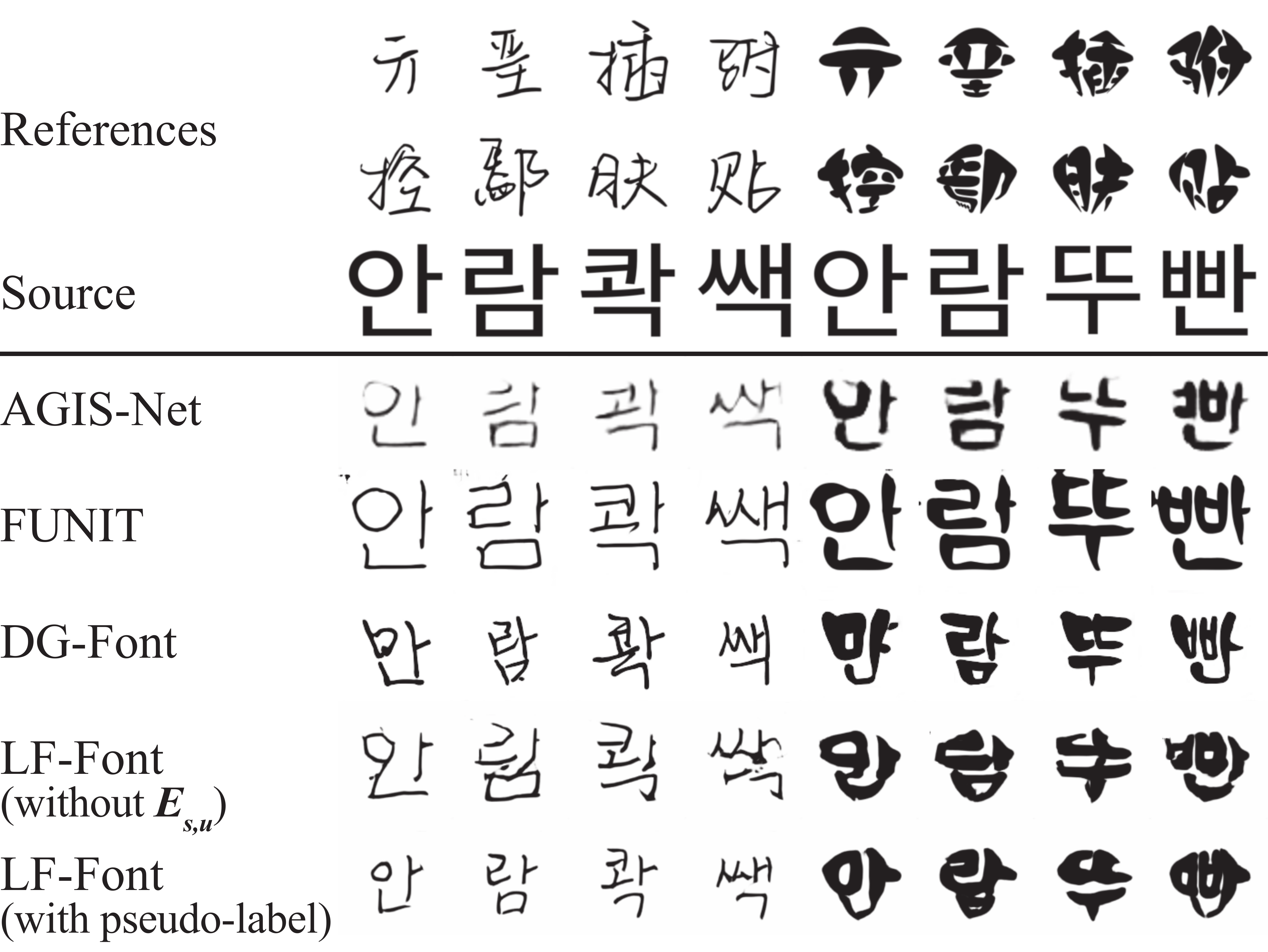}%
    \caption{\small {\bf Generation results on cross-lingual zero-shot scenario.} In this scenario, we generate Korean glyph images from Chinese references with models trained on Chinese script.}%
    \label{fig:zeroshot_examples}
\end{figure}

\begin{table}[ht!]
\small
\centering
\setlength{\tabcolsep}{4pt}
\addition{
\begin{tabular}{@{}lccc@{}}
\toprule
& \multicolumn{3}{c}{Accuracies $\uparrow$} \\ 
& S & C & H\\\midrule
AGIS-Net &	14.1 &	34.0 &	20.0 \\
DG-Font	& 46.5 &	41.2 &	43.6 \\
FUNIT	& 11.8 &	67.2 &	20.1 \\
\ours without $E_{s,u}$	& 52.1 &	35.2 &	42.0 \\
\ours with pseudo character label & 70.9 &	42.9 &	53.4 \\ \bottomrule
\end{tabular}
}
\caption{\addition{
\small {\bf Zero-shot generation results.} We train font generation models in Chinese and test them on the unseen Korean target characters with Chinese references.
}}
\label{table:zeroshot_results}
\end{table}

\myparagraph{Beyond font generation tasks.}
In this paper, we focus on few-shot font generation tasks by capturing complex local styles of the font domain. Although, our key idea on localized style representations can be extended to general generation tasks such as attributed-conditioned generation tasks, there are some challenges on extending \ours to general attributed-conditioned generation tasks.

As we discussed in \S\ref{sec:relwork}, there are two significant differences between font generation tasks and general attribute-conditioned generation tasks (\eg, attribute-conditional facial image generation \cite{yan2016attribute2image, choe2017face, lu2018attribute, he2019attgan}). First, a glyph is uniquely defined for each style, while an image can be diversely mapped to different images even with the same identity. Second, it is easy to collect a glyph image with the same content but different styles. Our method heavily relies on these two font-specific properties in the training phase. Especially, in the phase 1 training, we construct a mini-batch where the input images and the target image have a coherent style, where all components of the target image can be found in the input components (Figure \ref{fig:train_batch}). Our mini-batch construction strategy let the model learn component-wise representations but at the same time, limits the flexibility of \ours to general image domains. Note that in other visual domains (\eg, facial images) constructing a mini-batch with a coherent style (\eg, the same identity) but different attributes (\eg, different gender or different hair color) are difficult, so that many attributed-conditioned generation methods formulate unpaired image generation problems while we formulate font generation as the paired scenario. Extending \ours to unpaired image generation tasks and general image domains will be an interesting future research direction.

\section{Conclusion}

Our novel few-shot font generation method, named \ours, produces high-quality, complex glyphs using only a few reference images. By utilizing compositionality, a language-specific characteristics, \ours learns to capture local component-wise style representations from the given glyphs with the weak supervision of compositionally. That is, we only utilize the component labels, not the location of each component, skeleton, or stroke. To reduce the number of required references, we propose factorization modules, which derive the factors and then reconstruct the entire character-wise style representations from a few reference images. These factorization modules are trained on the seen character-wise style representations; but are well generalized to the unseen character-wise style representations by disentangling character-wise style representations into style and component factors. As a result, \ours can handle the characters unseen in the reference set, which is an important success factor when dealing with only a few reference glyphs. In the experiments, \ours outperformed six state-of-the-art few-shot font generation methods on both Chinese and Korean in various evaluation metrics, particularly in {\it style-aware} benchmarks. Our extensive analysis of our design choice supports the notion that our framework effectively disentangles content and style representations, resulting in high-quality generated samples with only a few references. Furthermore, to mitigate the inherent drawback of \ours, we employ an auxiliary character classifier on test-time. The proposed prediction-based inference strategy enables \ours to generate unseen language systems without losing localized style representations. Furthermore, we demonstrate that \ours can be used for data augmentation of character recognition systems.

\ifCLASSOPTIONcompsoc
  \section*{Acknowledgments}
  This research was supported by the Basic Science Research Program through the NRF Korea funded by the MSIP (NRF-2019R1A2C2006123, 2020R1A4A1016619), the IITP grant funded by the MSIT (2020-0-01361, Artificial Intelligence Graduate School Program (YONSEI UNIVERSITY), No.2021-0-02068 (Artificial Intelligence Innovation Hub)), and the Korea Medical Device Development Fund grant funded by the Korean government (Project Number:  202011D06).
\else
  \section*{Acknowledgment}
\fi

\ifCLASSOPTIONcaptionsoff
  \newpage
\fi

{\small
\bibliographystyle{IEEEtran}
\bibliography{references}
}

\begin{IEEEbiography}[{\includegraphics[width=1in,height=1.15in,clip,keepaspectratio]{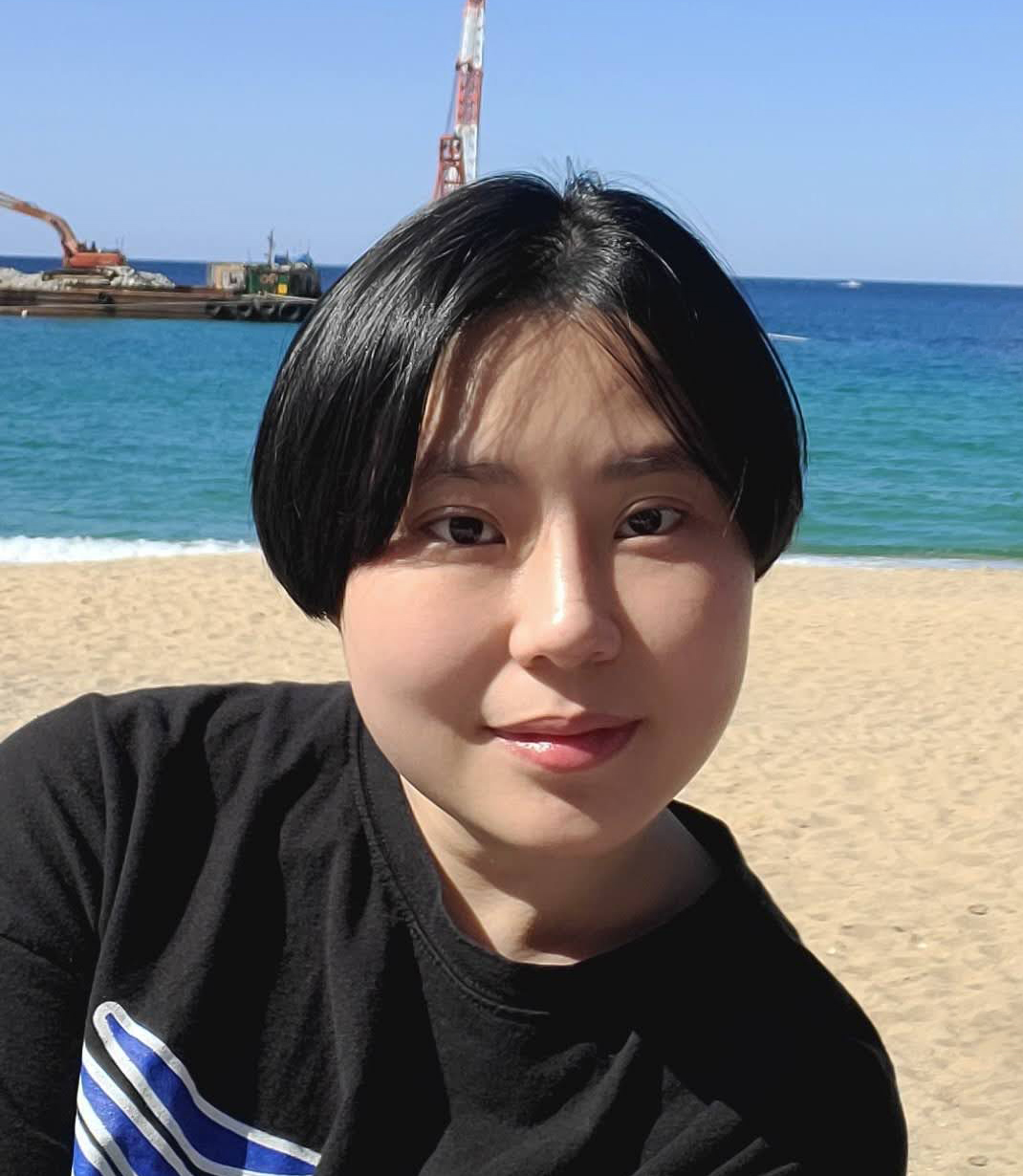}}]{Song Park} is also a Ph.D. candidate at the School of Integrated Technology, Yonsei University, South Korea. She received her B.S. degree in Integrated Technology from Yonsei University, Seoul, Korea, in 2016. Her recent research interests include computer vision and machine learning.
\end{IEEEbiography}

\begin{IEEEbiography}[{\includegraphics[width=1in,height=1.15in,clip,keepaspectratio]{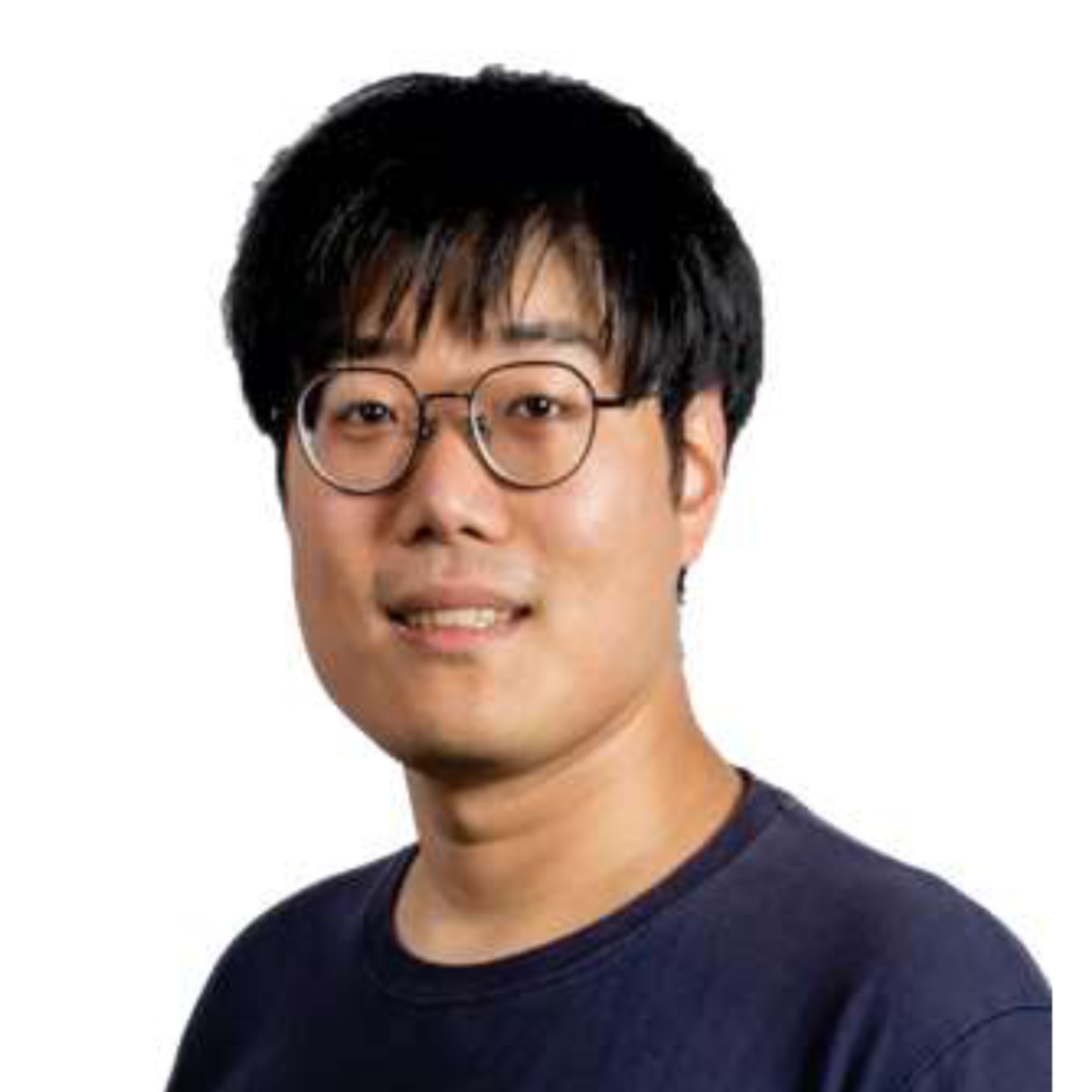}}]{Sanghyuk Chun} is a lead research scientist at the NAVER AI Lab. He was a research engineer at an advanced recommendation team in Kakao Corp from 2016 to 2018. He received his Master's and Bachelor's degrees in Electronical Engineering from KAIST, Daejeon, Korea, in 2016 and 2014, respectively. His research interests focus on reliable machine learning and vision-and-language.
\end{IEEEbiography}

\begin{IEEEbiography}[{\includegraphics[width=1in,height=1.15in,clip,keepaspectratio]{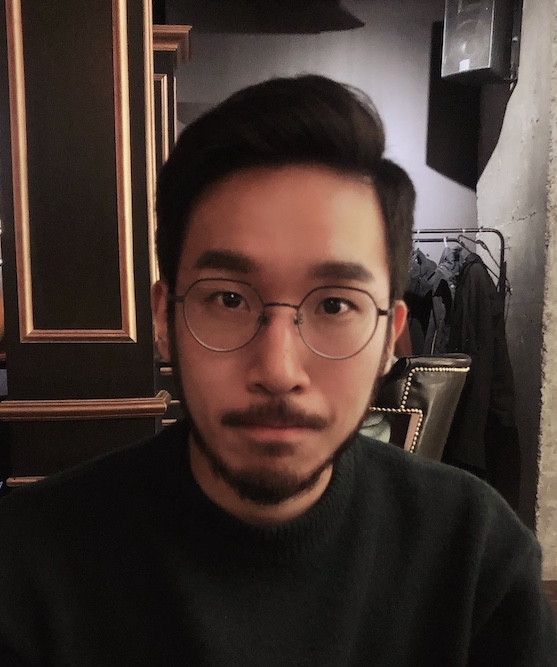}}]{Junbum Cha} is a research engineer at Clova OCR, NAVER Corp. He developed an AI go engine at the game AI team in the NHN Corp from 2017 to 2019. He received his Master's and Bachelor's degrees in computer science from Yonsei University, Seoul, Korea, in 2017 and 2015, respectively. His research interests include generative models, automated machine learning, and robust machine learning.
\end{IEEEbiography}

\begin{IEEEbiography}[{\includegraphics[width=1in,height=1.15in,clip,keepaspectratio]{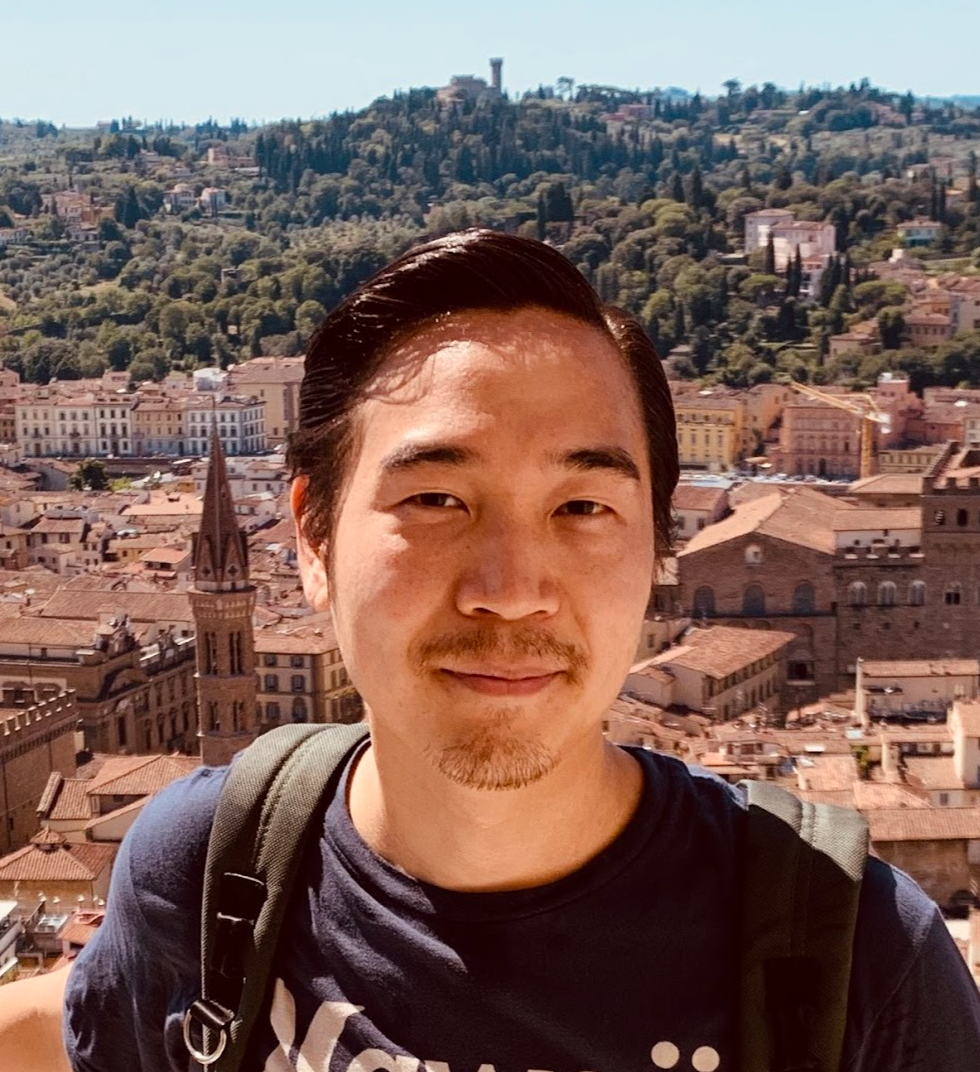}}]{Bado Lee} is a team leader and  developer at Clova OCR, NAVER Corp. He was a developer at the Samsung Mobile Division from 2012 to 2017. He received his Master's and Bachelor's degrees in Electrical Engineering from Seoul National University, Seoul, Korea, in 2012 and 2010, respectively. His research interests focus on computer vision and image processing.
\end{IEEEbiography}

\begin{IEEEbiography}[{\includegraphics[width=1in,height=1.25in,clip,keepaspectratio]{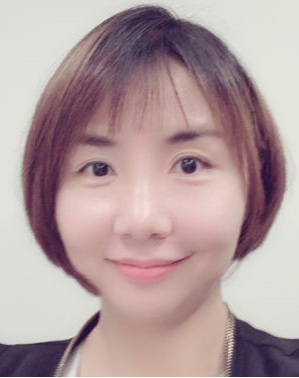}}]{Hyunjung Shim} received her B.S. in Electrical Engineering from Yonsei University, Seoul, Korea, in 2002, and her M.S. and Ph.D. in Electrical and Computer Engineering from Carnegie Mellon University, Pittsburgh, PA, USA, in 2004 and 2008, respectively. 
She was with Samsung Advanced Institute of Technology, Samsung Electronics Company, Ltd., Suwon, Korea, from 2008 to 2013. She is currently an associate professor at the School of Integrated Technology, Yonsei University. Her research interests include generative models, deep neural networks, classification/recognition algorithms, 3-D vision, inverse rendering, face modeling, and medical image analysis.
\end{IEEEbiography}

\begin{figure*}[ht]
    \centering
    \includegraphics[width=\linewidth]{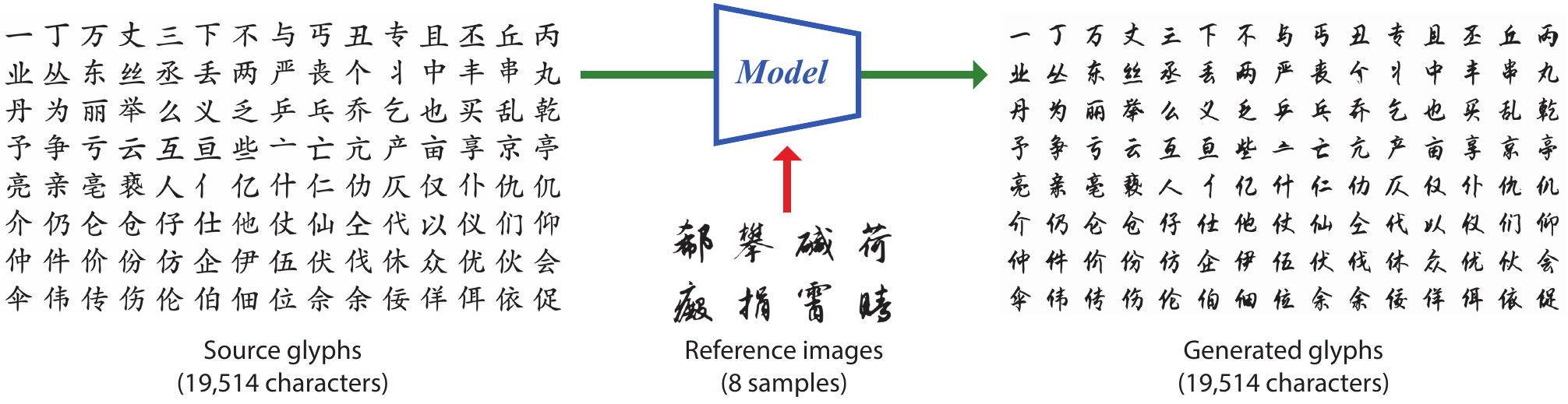}
    \caption{
    \small {\bf Overview of few-shot font generation tasks.} The few-shot font generation task aims to generate a full font library (19,514 characters in our Chinese generation scenario) with a coherent style with only a few references (eight glyphs in our experiments).
    }
    \label{fig:ffg_scenario}
\end{figure*}
\,
\begin{figure}[ht]
    \centering
    \includegraphics[width=\linewidth]{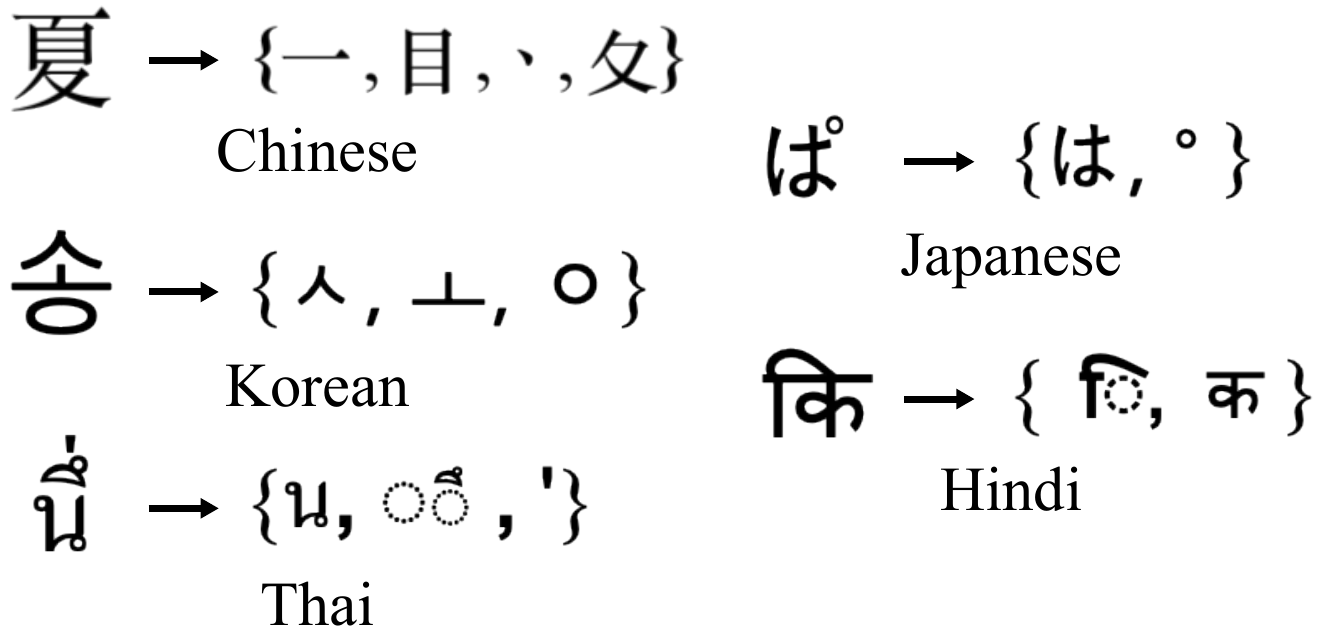}%
    \caption{\small {\bf Example compositionality of widely-used languages.}}%
    \label{fig:compositionality}
\end{figure}
\,
\begin{figure}[ht]
    \centering
    \includegraphics[width=\linewidth]{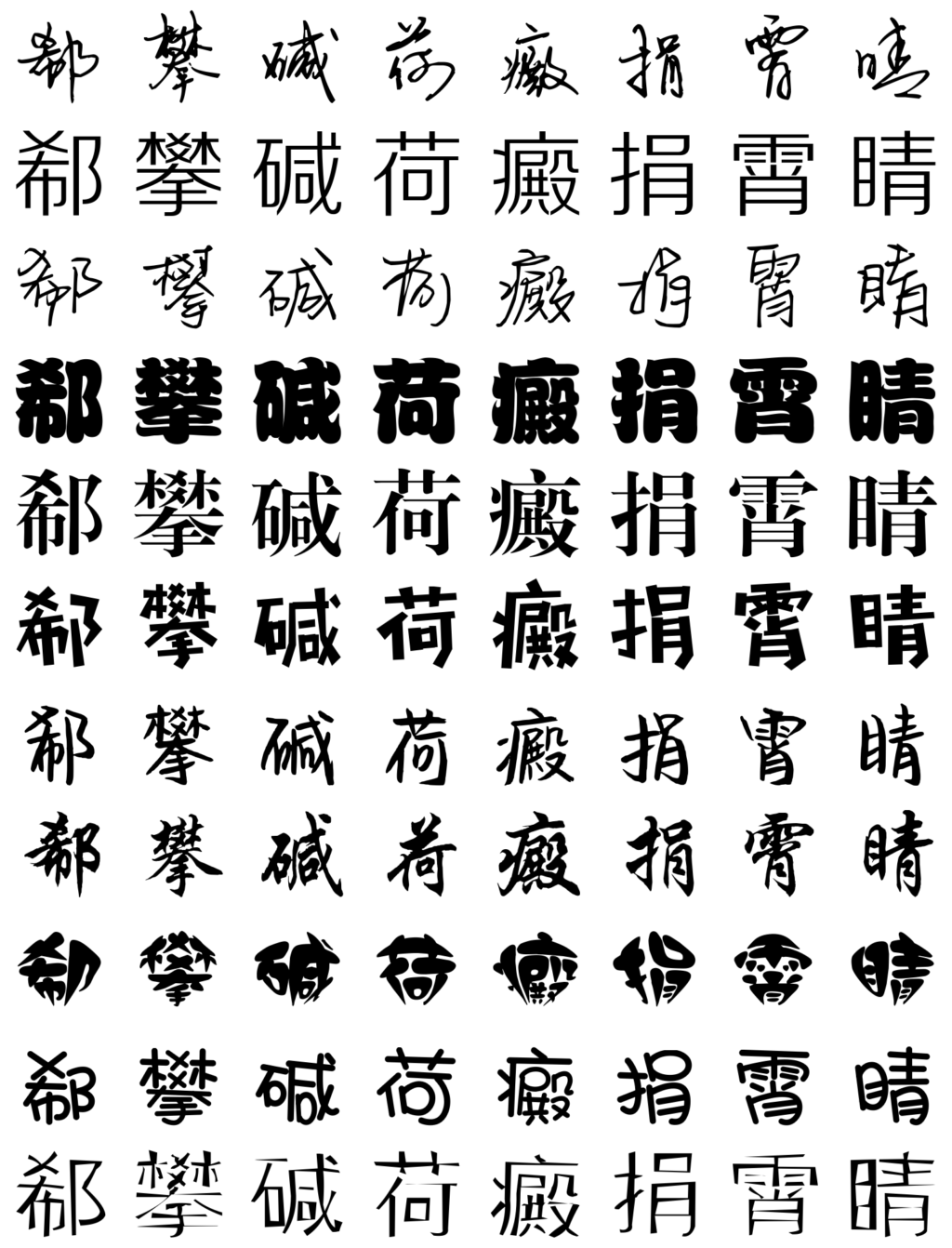}%
    \caption{
    \small {\bf Reference images with target styles.} We visualized the eight reference samples per style used in Figure 8. Each row corresponds to the two columns of Figure 8 in the same order.
    }%
    \label{fig:style_refs}
\end{figure}

\begin{figure*}
    \centering
    \includegraphics[width=\linewidth]{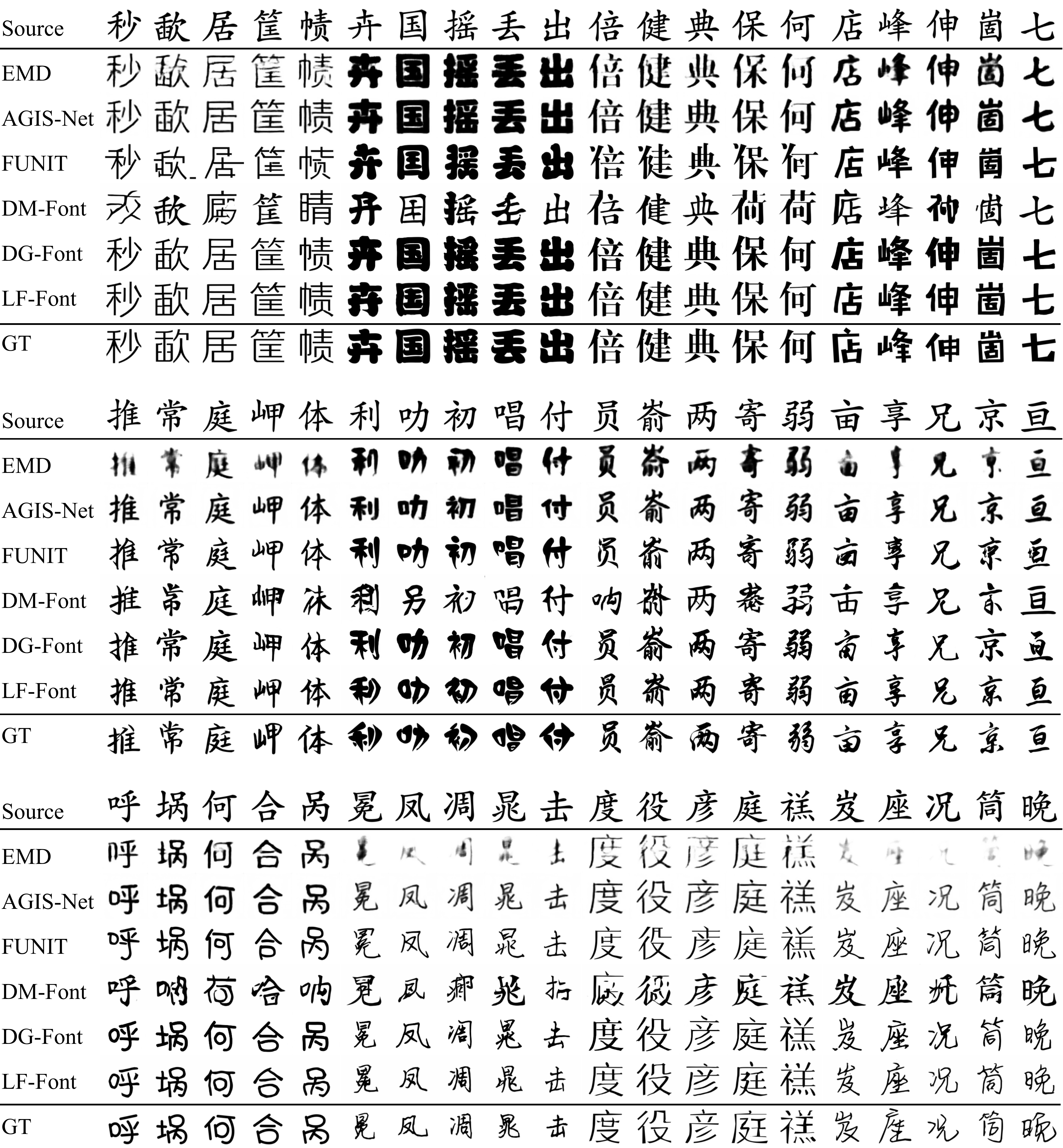}
    \caption{\small {\bf More generation samples.} We provide more generated glyphs.}
    \label{fig:more_results}
\end{figure*}

\end{document}